\pdfoutput=1
\documentclass[11pt, letterpaper, shortlabels]{berkeley}

\usepackage{hyperref}
\usepackage{color-edits}

\newcommand{\eg}{{e.g.}}
\newcommand{\etc}{{e.t.c.}}
\newcommand{\ie}{{i.e.}} 

\ifx\assumption\undefined
\newtheorem{assumption}{Assumption}
\fi

\usepackage[capitalize,noabbrev]{cleveref}
\makeatletter
\def\adl@drawiv#1#2#3{%
        \hskip.5\tabcolsep
        \xleaders#3{#2.5\@tempdimb #1{1}#2.5\@tempdimb}%
                #2\z@ plus1fil minus1fil\relax
        \hskip.5\tabcolsep}
\newcommand{\cdashlinelr}[1]{%
  \noalign{\vskip\aboverulesep
           \global\let\@dashdrawstore\adl@draw
           \global\let\adl@draw\adl@drawiv}
  \cdashline{#1}
  \noalign{\global\let\adl@draw\@dashdrawstore
           \vskip\belowrulesep}}
\makeatother
\newcommand{\method}{CoSER\xspace}

\usepackage{wrapfig}
\title{\method: A Comprehensive Literary Dataset and Framework for Training and Evaluating LLM Role-Playing and Persona Simulation}

\usepackage[all]{hypcap}

\usepackage[authoryear, round]{natbib}

\usepackage{hyperref}[citecolor=magenta,linkcolor=magenta]

\hypersetup{
    colorlinks = true,
    citecolor = {magenta},
}

\usepackage{multirow, makecell, caption}\usepackage{microtype}
\usepackage{graphicx}
\usepackage{booktabs} 

\usepackage{amsmath}
\usepackage{amssymb}
\usepackage{mathtools}
\usepackage{amsthm}
\usepackage{mathrsfs}
\usepackage{nicefrac}
\usepackage{dsfont}
\usepackage{enumitem}
\usepackage{ulem}
\usepackage{arydshln}
\usepackage{xspace}
\usepackage[capitalize,noabbrev]{cleveref}
\usepackage{subcaption}
\usepackage{wrapfig}
\usepackage{lipsum}
\usepackage{listings}

\usepackage{amsmath}
\usepackage{amssymb}
\usepackage{mathtools}
\usepackage{amsthm}
\usepackage{bbm}

\usepackage{algpseudocode}
\usepackage{setspace}

\usepackage{color}

\usepackage{tabularx}
\newcolumntype{C}{>{\centering\arraybackslash}X}

\definecolor{mygray}{RGB}{226, 226, 226}
\definecolor{myred}{RGB}{252, 142, 142}
\definecolor{mygreen}{RGB}{147, 255, 143}
\definecolor{myblue}{RGB}{144, 155, 255}
\definecolor{myyellow}{RGB}{253, 253, 143}
\definecolor{mypurple}{RGB}{255, 142, 250}

\newcommand{\dq}[1]{``#1''}

\newcommand{\persona}{c}
\newcommand{\personadata}{\mathcal{D}}
\newcommand{\agent}{\pi}

\usepackage[export]{adjustbox}

\newcommand\pythonstyle{\lstset{
basicstyle=\ttfamily\footnotesize,
language=Python,
morekeywords={self, clip, exp, mse_loss, uniform_sample, concatenate, logsumexp},              
keywordstyle=\color{deepblue},
emph={MyClass,__init__},          
emphstyle=\color{deepred},    
stringstyle=\color{deepgreen},
frame=single,                         
showstringspaces=false
}}

\lstnewenvironment{python}[1][]
{
\pythonstyle
\lstset{#1}
}
{}

\newcommand\pythoninline[1]{{\pythonstyle\lstinline!#1!}}

\makeatletter
\def\mathcolor#1#{\@mathcolor{#1}}
\def\@mathcolor#1#2#3{%
  \protect\leavevmode
  \begingroup
    \color#1{#2}#3%
  \endgroup
}
\makeatother

\Crefformat{equation}{#2Eq.\;(#1)#3}

\Crefformat{figure}{#2Figure #1#3}
\Crefformat{assumption}{#2Assumption #1#3}
\Crefname{assumption}{Assumption}{Assumptions}

\usepackage{crossreftools}
\usepackage{dsfont}
\usepackage{nicefrac}

\author[1,2]{Xintao Wang}
\author[2]{Heng Wang}
\author[1,2]{Yifei Zhang}
\author[1]{Xinfeng Yuan}
\author[1]{Rui Xu}
\author[3]{Jen-tse Huang}
\author[1]{Siyu Yuan}
\author[1]{Haoran Guo}
\author[1]{Jiangjie Chen}
\author[2]{Shuchang Zhou}
\author[1]{Wei Wang}
\author[1]{Yanghua Xiao}

\affil[1]{Fudan University}
\affil[2]{StepFun}
\affil[3]{Johns Hopkins University}

\correspondingauthor{Corresponding author(s): xtwang21@m.fudan.edu.cn. Work done during Xintao Wang's internship at StepFun.}

\begin{abstract}
\textbf{Abstract:} Role-playing language agents (RPLAs) have emerged as promising applications of large language models (LLMs).
However, simulating established characters presents a challenging task for RPLAs, due to the lack of authentic character datasets and nuanced evaluation methods using such data. 
In this paper, we present \method, a collection of a high-quality dataset, open models, and an evaluation protocol towards effective RPLAs of established characters.  
The \method dataset covers 17,966 characters from 771 renowned books.
It provides authentic dialogues with real-world intricacies, as well as diverse data types such as conversation setups, character experiences and internal thoughts. 
Drawing from acting methodology, we introduce given-circumstance acting for training and evaluating role-playing LLMs, where LLMs sequentially portray multiple characters in book scenes.  
Using our dataset, we develop \method 8B and \method 70B, \ie, advanced open role-playing LLMs built on LLaMA-3.1 models.  
Extensive experiments demonstrate the value of the \method dataset for RPLA training, evaluation and retrieval. 
Moreover, \method 70B exhibits state-of-the-art performance surpassing or matching GPT-4o on our evaluation and three existing benchmarks, \ie, 
achieving 75.80\% and 93.47\% accuracy on the InCharacter and LifeChoice benchmarks respectively. 
Our code, dataset and models are available at:  \href{https://github.com/Neph0s/CoSER}{https://github.com/Neph0s/CoSER}.

\end{abstract}

\begin{document}

\maketitle
\begin{figure}[!t]
    \centering
    \includegraphics[width=0.6\linewidth]{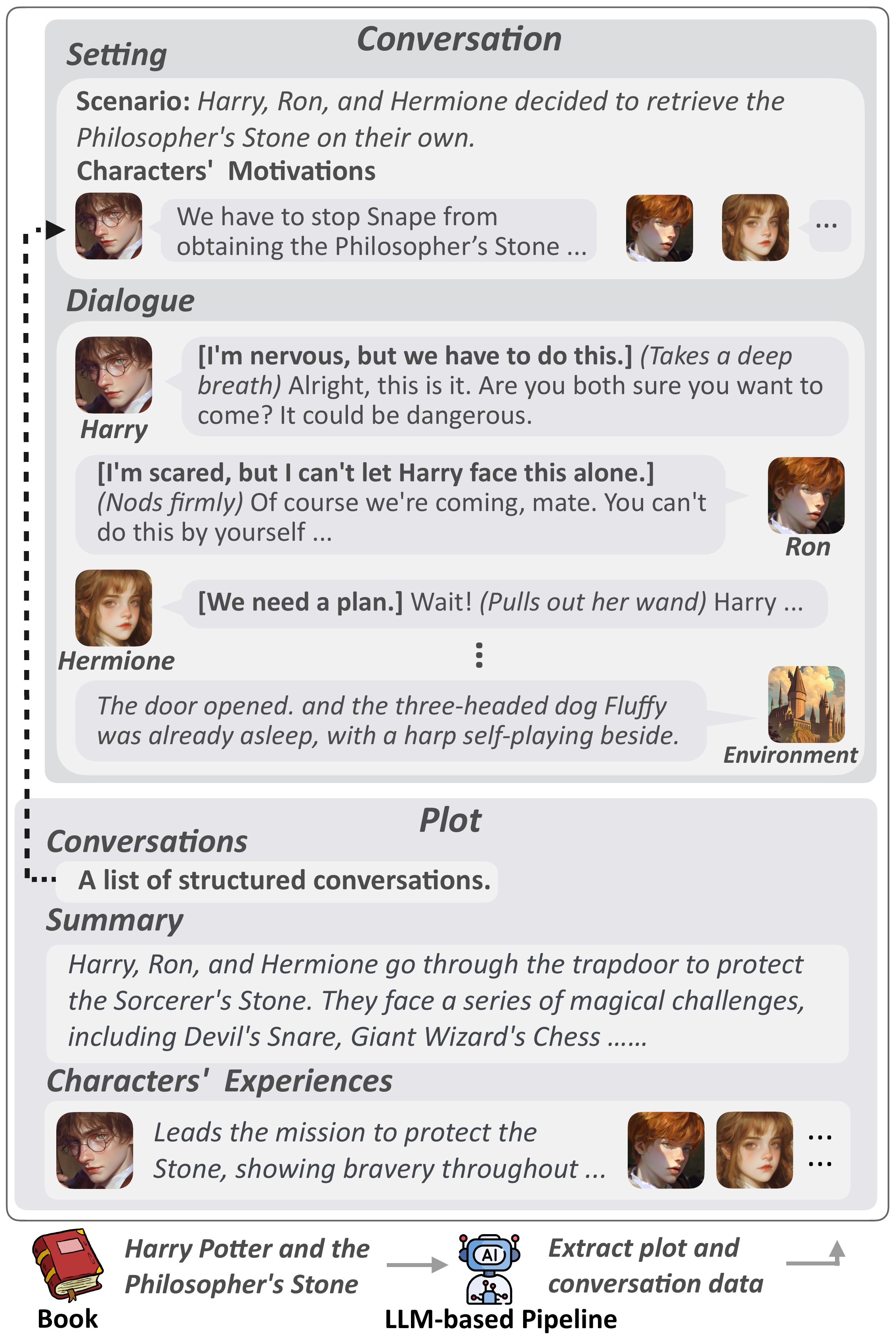}
    \caption{
    An example from \method dataset, which provides 
    comprehensive data types 
    such as conversation dialogues and settings, plot summaries, characters' inner thoughts, authentically sourced from renowned books.
    }
    \label{fig:front}
\end{figure}

\section{Introduction}
Recent advances in large language models (LLMs) have facilitated the emergence of anthropomorphic cognition in AI~\citep{kosinski2023theory, shanahan2023role}.
Role-playing language agents (RPLAs), \ie, LLMs that simulate specific personas based on relevant data, have hence been popular~\citep{Park2023GenerativeAgents}.
RPLAs have been adopted to simulate personas of various types, including demographics, characters, or daily individuals~\citep{chen2024from}.
They have inspired extensive applications including character chatbots, agents in video games, and digital clones for humans. 
This paper studies \textbf{RPLAs for established characters}, which represent a crucial yet challenging task beyond the naive portrayal of individual traits or stereotypes. 
Specifically, RPLAs should faithfully align with their characters' complex backgrounds and capture their nuanced personalities.

Towards effective RPLAs, two major challenges persist in:
\textit{1)} \textbf{\textit{Data}}: High-quality datasets are lacking.
Existing datasets are limited to dialogues between two characters, and lack necessary dialogue contexts and knowledge in other forms.
Moreover, many datasets are synthesized by LLMs, compromising authenticity and fidelity to the origins~\citep{wang2023rolellm, lu2024large};
\textit{2)}  \textbf{\textit{Evaluation}}: Current methods fall short in assessing complex character portrayals of LLMs. 
They typically focus on single-turn interactions with pre-defined question sets, and rely on either LLM-based judges or multi-choice questions. 
The former lack nuanced discrimination and suffer from bias issues~\citep{li2024judgesurvey}, while the latter only assess specialized aspects~\citep{xu2024character}.
Overall, there is a lack of authentic character data and appropriate evaluation methods based on such data.

In this paper, we introduce \method, a collection of authentic character data, along with open state-of-the-art models and evaluation protocol based on such data, for \textbf{Co}ordinating LLM-Based Persona \textbf{S}imulation of \textbf{E}stablished \textbf{R}oles. 
The \method dataset is sourced from narratives and dialogues in 771 renowned books, processed via our LLM-based pipeline. 
\method differs from existing datasets in two fundamental ways:  
\textit{1)} \method extracts authentic, multi-character dialogues from acclaimed literary works, 
in contrast to LLM-synthesized question-answer pairs in previous work. 
Hence, our dataset maintains high source fidelity while exhibiting greater quality and complexity. 
\textit{2)}  \method incorporates comprehensive types of data, as shown in Fig. ~\ref{fig:front}:  
\textit{i)}  Besides character profiles and dialogues, \method encompasses plot summaries, character experiences, and conversation backgrounds, supporting various purposes including prompting, retrieval, model training and evaluation. 
\textit{ii)}  Conversations in \method capture characters' actions and internal thoughts beyond surface-level speech, enabling RPLAs to simulate sophisticated cognitive and behavioral processes of humans, such as \dq{\textit{[I'm nervous, but we have to do this] (Takes a deep breath) Alright, we ...}}. 
We provide a clear comparison between \method and existing datasets in Table ~\ref{tab:dataset_overview}.

We introduce given-circumstance acting (GCA) for training and evaluating role-playing LLMs, leveraging \method dataset. 
Given a conversation with messages $M$, characters $\mathcal{C}$ and setting $\mathcal{S}$, GCA requires an actor LLM to sequentially portray each character $c\in\mathcal{C}$ to recreate the conversation, as illustrated in Fig. ~\ref{fig:main}.
During training, we train LLMs to portray each character $c$, on their authentic utterances $M_c\subset M$. 
As a result, we develop \method 8B and 70B, built on LLaMA-3.1 models~\citep{dubey2024llama}, which demonstrate true-to-life character portrayal and state-of-the-art performance on multiple RPLA benchmarks. 
For evaluation, GCA involves two steps: multi-agent simulation and penalty-based LLM judging.
Given a test conversation $M$, we:
\textit{1)}  create a multi-agent system to simulate a  conversation $\bar{M}$, where the actor LLM portrays each character $c
\in\mathcal{C}$ in the same setting $\mathcal{S}$; 
\textit{2)}  assess $\bar{M}$ using penalty-based LLM critics, leveraging detailed rubrics and the original conversation $M$.  
GCA evaluation offers three advantages: 
First, it comprehensively reflects actor LLMs' abilities via multi-agent simulation;  
Second, it is based on authentic scenes and groundtruth dialogues. 
Third, it provides expert-curated rubrics to guide LLM critics.

Our contributions are summarized as follows:
\begin{enumerate} 
    \item We introduce the \method dataset and models for RPLA research and applications. Our dataset comprises 29,798 authentic conversations and comprehensive types of data from 771 renowned books. Leveraging this dataset, we develop \method 8B and \method 70B, which are state-of-the-art models for RPLAs. 
    \item We propose given-circumstance acting for training and evaluating role-playing LLMs, drawing from established acting theory. Our evaluation comprehensively tests actor LLMs via multi-character simulation, while providing original dialogues and detailed rubrics to enhance LLM-based assessment. 
    \item Results of extensive experiments demonstrate the significant value of our dataset for the training, retrieval and evaluation of RPLAs. Notably, the \method models achieve state-of-the-art performance across four benchmarks and human evaluation for RPLAs. 
\end{enumerate}

\section{Related Work}

\begin{table*}[t]
    \centering
    \setlength{\tabcolsep}{4pt}
    \renewcommand{\arraystretch}{0.9}
    \resizebox{\textwidth}{!}{%
    \begin{tabular}{@{}lccccccccccccc@{}}
    \toprule
            \multirow{2}{*}{\textbf{Dataset}} & \textbf{Book} & \multicolumn{3}{c}{\textbf{Character}} & \multicolumn{5}{c}{\textbf{Conversation}} & \multicolumn{3}{c}{\textbf{Message}} & \textbf{Plot} \\
            \cmidrule(lr){2-2} \cmidrule(lr){3-5} \cmidrule(lr){6-10} \cmidrule(lr){11-13} \cmidrule(r){14-14} 
            & \textbf{Num.} & \textbf{Num.} & \textbf{Profile} & \textbf{Expr.} & \textbf{\#Conv.} & \textbf{\#Turns} & \textbf{Setting} & \textbf{Auth.} & \textbf{Multi-Chara.} & \textbf{Speech} & \textbf{Thought} & \textbf{Action} & \textbf{Summ.} \\ \midrule 
    Charater-LLM & & 9 & \checkmark & & 14,300 & 13.2 & \checkmark & & & \checkmark & & & \\
    ChatHaruhi & & 32 & \checkmark & & 54,726 & $>$2 & & \checkmark* & \checkmark & \checkmark & & & \\
    RoleLLM & & 100 & \checkmark & & 140,726 & 2 & & & & \checkmark & & & \\
    HPD & 7 & 113 &  & & 1,191 & 13.2 & \checkmark & \checkmark & \checkmark & \checkmark & & & \checkmark \\
    LifeChoice &  388 & 1,462 & \checkmark & & 1,462 & 2 & \checkmark & \checkmark & & & & & \\
    CroSS-MR & 126 & 126 & \checkmark & & 445 & 2 & \checkmark & \checkmark & &  & & & \\
    CharacterGLM & & 250 & \checkmark & & 1,034 & 15.8 & \checkmark & & & \checkmark & & & \\
    CharacterEval & & 77 & \checkmark & & 1,785 & 9.3 & \checkmark &  \checkmark & & \checkmark &  & \checkmark & \\
    DITTO & & 4,002 & \checkmark & & 7,186 & 5.1 & & & & \checkmark & & & \\
    MMRole & & 85 & \checkmark & & 14,346 & 4.2 & & & & \checkmark & & & \\
    CharacterBench & &3,956 & \checkmark & & 13,162 & 11.3 & & & & \checkmark & & & \\
    \method & 771 & 17,966 & \checkmark & \checkmark & 29,798 & 13.2 & \checkmark & \checkmark & \checkmark & \checkmark & \checkmark & \checkmark & \checkmark\\
    
    \bottomrule
    \end{tabular}%
    }
    \caption{Overview of \method and existing RPLA datasets. 
    For characters, Num. count characters with profiles, and Expr. denotes structured character experiences. 
    For conversations, Auth. indicates authentic dialogues or behaviors from the books, and Multi-Chara. denotes involving more than 2 characters. Num. (number), Conv. (conversation), and Summ. (summary) are abbreviations. 
    }
    \label{tab:dataset_stats}
\end{table*}

Using LLMs to simulate human personas represents a pioneering research direction. 
Early studies develop prototypes of RPLAs for fictional characters~\citep{shao2023character} and multi-agent systems that simulate human society~\citep{Park2023GenerativeAgents}, while exploring the nature~\citep{shanahan2023role} and potential limitations~\citep{liu-etal-2024-evaluating-large, deshpande-etal-2023-toxicity} of LLM role-play. 
~\citet{chen2024from} present a comprehensive survey of relevant research.

Specifically, an RPLA leverages LLMs to create a simulated persona $\agent_\persona$ that simulates a real character $\persona$ based on its persona data $\personadata_\persona$. 
Effective RPLAs require both comprehensive, high-quality data $\personadata_\persona$ and advanced role-playing LLMs. 
Among various persona types, we focus on RPLAs for established characters, which should faithfully align with their characters’ complex backgrounds and nuanced personalities.

\textbf{Datasets for RPLAs} \quad 
Persona data $\personadata_\persona$ describe the real persona $\persona$ through various representations, including profiles~\citep{yuan2024evaluating}, dialogues~\citep{wang2023rolellm}, experiences~\citep{li2023chatharuhi} and multimodal information~\citep{dai2024mmrolecomprehensiveframeworkdeveloping}, \etc. 
As shown in Table ~\ref{tab:dataset_overview}, existing datasets have several limitations. 
\textit{1)} Many are synthesized via LLMs' responses to general instruction sets~\citep{wang2023rolellm, chan2024personahub} or character-specific questions~\citep{shao2023character}, such as RoleBench~\citep{wang2023rolellm}.  
However, LLM-synthesized data compromise authenticity and fidelity to original sources.
\textit{2)} Human-annotated datasets such as CharacterEval~\citep{tu2024charactereval} and CharacterDial~\citep{zhou2023characterglm} offer improved quality, but are expensive and difficult to scale.
\textit{3)} Several efforts extract authentic dialogues from fictional works, such as ChatHaruhi~\citep{li2023chatharuhi} and HPD~\citep{chen2023large}. However, they rely on human efforts for individual sources and are hence hard to scale as well.
\textit{4)} Furthermore, existing datasets offer limited representations and forms, \ie, mainly consisting of two-character or user-character question-answer pairs.
These datasets support various purposes, including prompting, training, retrieval augmentation, and evaluation of RPLAs.

\textbf{Evaluation for RPLAs} \quad
Existing evaluation methods are based on either LLM judges or multi-choice questions~\citep{chen2024from}. 
LLM-judged methods typically elicit LLMs’ role-playing performance via predefined questions, and score the performance using LLM judges or reward models~\citep{chen2024from}.
They assess various dimensions, including character-independent aspects such as anthropomorphism~\citep{tu2024charactereval} and attractiveness~\citep{zhou2023characterglm}, as well as character-specific traits such as language style, knowledge, and personality~\citep{wang2023rolellm, shao2023character}. 
However, LLM judges suffer from inherent biases, \eg, length and position bias~\citep{li2024judgesurvey}, and may lack the necessary knowledge for character-specific evaluation.
Other benchmarks evaluate role-playing LLMs through multiple-choice questions, assessing specific aspects such as knowledge~\citep{shen2023roleeval}, decision-making~\citep{xu2024character}, motivation recognition~\citep{yuan2024evaluating}, and personality fidelity~\citep{wang2024incharacter}.

\section{\method Dataset}

In this section, we introduce the \method dataset, which covers authentic data of 17,966 characters from 771 renowned books. 
\method features its authentic, non-synthesized dialogues with real-world intricacies, and comprehensive data representations supporting various usages. 
In Table ~\ref{tab:dataset_stats}, we provide a comprehensive comparison with existing datasets. 
We illustrate our dataset's design principles in \S\ref{sec:data_design},  curation pipeline in \S\ref{sec:data_pipeline}, and statistical analysis in \S\ref{sec:data_statistics}.

\begin{figure*}[!t]
    \centering
    \includegraphics[width=\textwidth, center]{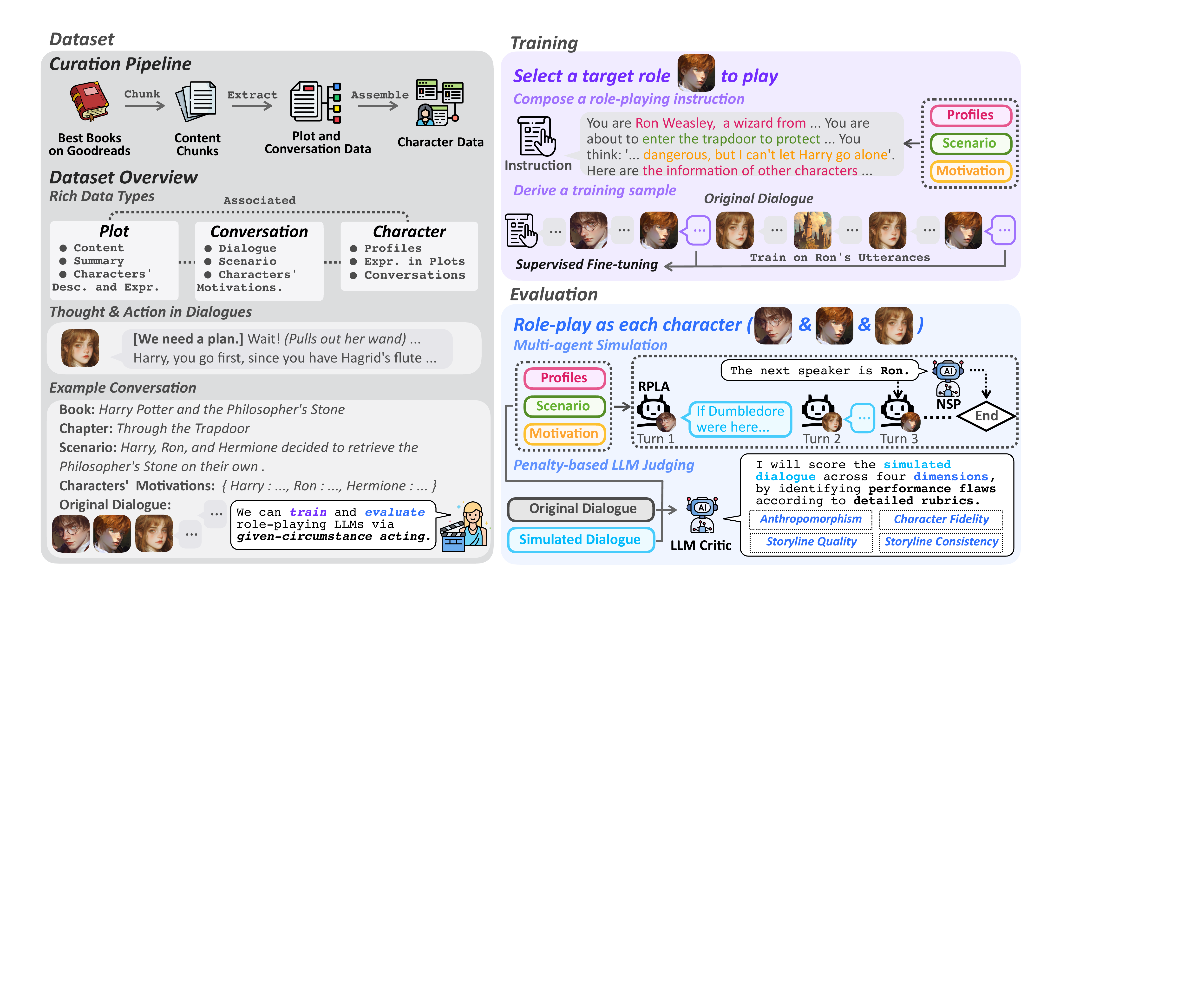}
    \vspace{0.2cm}
    \caption{
    Overview of \method's dataset, training and evaluation. 
    Left: The \method dataset is sourced from renowned books and processed via LLM-based pipeline. 
    It contains rich data types on plots, conversations and characters.  
    Right: 
    We apply given-circumstance acting to train and evaluate role-playing LLMs using these conversations.  
    For training, each sample trains the LLM to portray  a specific character in a conversation, using their  original dialogue.  
    For evaluation, 
    we build a multi-agent system for conversation simulation given the same scenario, and assess the simulated dialogue via  penalty-based LLM critics. 
    }
    \label{fig:main}
\end{figure*}

\subsection{Design Principles}
\label{sec:data_design}

As shown in Table ~\ref{tab:dataset_stats}, \method differs from previous RPLA datasets mainly in its:  
\textit{1)} rich data types, 
\textit{2)} internal thoughts and physical actions in messages,
\textit{3)} environment as a role.

\textbf{Rich Types of Data} \quad 
The persona data $\personadata_\persona$ can represent a character $\persona$ from fictional works in diverse forms, \eg, narratives, profiles, dialogues, experiences, \etc. 
Previous work focuses primarily on profiles and dialogues, which represent limited knowledge.  
Hence, we propose a more comprehensive set of data types that are: 
\textit{1)} Comprehensive: covering extensive knowledge about characters and plots from the books; 
\textit{2)} Orthogonal: carrying distinct, complementary information with little redundancy;
\textit{3)} Contextual-rich: providing sufficient context to enable $\agent_\persona$ to faithfully reproduce $\persona$'s behaviors and responses in given scenarios.

Specifically, we organizes knowledge from books hierarchically via three interconnected elements: plots, conversations and characters. 
Each \textbf{plot} comprises its raw text, summary, conversations in this plot, and key characters' current states and experiences in this plot. 
A \textbf{conversation} contains not only the dialogue transcripts, but also rich contextual settings including scenario descriptions and characters' motivations. 
\textbf{Characters} are associated with their conversations and plots, based on which we craft their profiles. 

\textbf{Thoughts and Actions in Messages} \quad 
Previous RPLA studies typically restrict RPLAs' output space to verbal speech alone, limiting their ability to fully represent human interactions. 
In this paper, we extend the message space of RPLAs and character datasets into three distinct dimensions: speech ($\mathcal{L}$), action ($\mathcal{A}$), and thought ($\mathcal{T}$), significantly enriching the expressiveness. 
For instance, an RPLA can convey silence by generating only thoughts and actions without verbal speech. 
The three dimensions are distinguished by markup symbols and function mechanisms:  
\begin{itemize}[itemsep=-3pt, topsep=0pt, partopsep=0pt]
    \item \textbf{Speech} is for verbal communications of characters.
    \item \textbf{Action} captures physical behaviors, body language, facial expressions, \etc. Similar to  tool use in agents~\citep{weng2023agent}, actions can be programmed to trigger downstream events in multi-agent systems. 
    \item \textbf{Thought} represents internal thinking processes, which enable RPLAs to simulate sophisticated human cognition. 
    Thoughts should be invisible to others, forming  information asymmetry~\citep{zhou2024sotopia}. 
\end{itemize}

\textbf{Environment as a Role} \quad 
In RPLA applications like AI TRPG~\footnote{Tabletop Role-Playing Games}~\citep{liang2023tachikuma}, LLMs often serve as world simulators that respond to players' actions. 
To promote this ability, we consider environment as a special role $e$, which provide environmental responses such as physical changes and reactions from unspecified characters or crowds.

\subsection{Dataset Curation} 
\label{sec:data_pipeline}

We curate the \method dataset through a systematic LLM-based pipeline that transforms book content into high-quality data for RPLAs  
~\footnote{In this paper, we employs Claude-3.5-Sonnet (20240620).}. 
The details are as follows. 

\textbf{Source Selection} \quad 
Our dataset is sourced from most acclaimed literary works to ensure data quality and character depth. 
We identify the top 1,000 books on \textit{Goodreads}'s \textit{Best Books Ever} list~\footnote{https://www.goodreads.com/list/show/1.Best\_Books\_Ever}, and obtain the content for 771 books.
As shown in Table~\ref{tab:selected_books}, these books  offer characters and narratives with literary significance and widespread recognition across diverse genres, time periods, and cultural backgrounds.

\textbf{Chunking} \quad 
We segment book contents into chunks to fit in LLMs' context window. 
We employs both static, chapter-based strategy and dynamic, plot-based strategy. 
Initially, we use regular expressions to identify chapter titles as natural chunk boundaries. 
Then, we merge adjacent small chunks and split large chunks to ensure moderate chunk sizes. 
However, static chunking neglects the storyline and truncates important plots or conversations. 
To address this, we implement dynamic plot-based chunking, \ie, during data extraction, we also prompt LLMs to identify truncated plots or trailing content in the current chunk, and concatenate them with the subsequent chunk to ensure plot integrity.

\textbf{Data Extraction} \quad 
We employ LLMs to extract plot and conversation data from book chunks, including (1) contents, summaries and character experiences of plots, and (2) dialogues and background settings of conversations. 
The extracted data representations are illustrated in Fig. ~\ref{fig:front} and introduced in \S\ref{sec:data_design}. 
In the messages, speeches are always extracted from the original dialogues, while actions and thoughts can either be extracted or inferred by LLMs based on the context. 
For evaluation purposes, we hold out data from the final 10\%  plots in each book.

\textbf{Organizing Character Data} \quad 
Based on the extracted data, we form the knowledge bases for characters in three steps.  
First, we unify character references by establishing name mappings between aliases and canonical names using LLMs, \eg, mapping \textit{Lord Snow} to an unified identifier \textit{Jon Snow}. 
Second, we aggregate relevant plots and conversations for each character. 
Finally, we leverage LLMs to generate character profiles based on their extracted data, describing them from multiple perspectives including background, experiences, physical characteristics, personality traits, core motivations, relationships, character arcs, \etc. 

For technical details, including our prompts, engineering implementation, and handling mechanisms for exception caused by LLMs, please refer to ~\S\ref{sec:app_dataset}.

\section{Training and Evaluation via GCA}
\label{sec:method}

In this section, we introduce \textit{given-circumstance acting} (GCA) for training and evaluating LLMs' role-playing abilities using the \method dataset, as shown in Fig. ~\ref{fig:main}.

\subsection{Given-Circumstance Acting}
\label{sec:gca}

In \textit{Konstantin Stanislavski}'s acting methodologies, given-circumstance acting is a fundamental approach where actors are trained and judged through performance within specified conditions including environmental context, historical events and personal conditions ~\citep{stanislavski2008actor}. 

We propose to adapt this approach to a framework that trains and evaluates LLMs' role-playing skills, leveraging the comprehensive data in \method.
In this framework, given a conversation with dialogue messages $M$, involved characters $\mathcal{C}$, and contextual setting $\mathcal{S}$, an actor LLM sequentially plays the role of each character $c\in\mathcal{C}$ to simulate the conversation. 

\subsection{GCA Training and \method Models}
\label{sec:training}

We fine-tune LLMs' role-playing abilities through GCA.  
Each training sample is derived from a conversation and one of its character $c$ in \method dataset, and LLMs are trained on $c$'s utterances $M_c$. 
Specifically, we first compose a role-playing instruction $i_\persona$ comprising the scenario description, the character's profile $p_\persona$ and motivation, and profiles of other involved characters,  which provide comprehensive context for role-playing. 
The original dialogue messages are denoted as $M=[m_1, ..., m_T]$, where $T$ is the number of turns. 
Then, the training sample $[i_\persona, m_1, ..., m_T]$ is a concatenation of the instruction and messages, where the character's messages $M_\persona\subset M$ are treated as outputs for optimization, and the other parts serve as inputs.  

We train \method 8B and \method 70B based on LLaMA 3.1 Instruct models~\citep{dubey2024llama}, using 90\% books in our dataset. 
To effectively support diverse use cases of RPLAs, 
our training samples cover extensive settings: 
\textit{1)} The \method dataset contains massive characters and conversation settings from extensive books.  
We train models on all characters in each conversation, ranging from major characters with detailed profiles to minor roles driven only by the context;  
\textit{2)} To simulate real use cases, we incorporate role-playing instructions in diverse formats through instruction templates of varying formats. 
Besides, we consider different combinations of available data by including or excluding: profiles of other characters, plot summaries, and characters' motivations;   
\textit{3)} We train models both with and without characters' internal thoughts in the extracted dialogues. 

We extend \method's training beyond character role-playing to develop complementary capabilities in environment modeling and next speaker prediction (NSP), which facilitates RPLA applications.  
To maintain models' general abilities, we augment our training data with the Tulu-3 dataset~\citep{lambert2024tulu}. 
Please refer to \S\ref{sec:app_training} for more details.

\subsection{GCA Evaluation}
\label{sec:eval}

Evaluating role-playing LLMs remains a significant challenge, primarily in two aspects: 
\textit{1)} providing appropriate scenarios to elicit role-playing performance, and 
\textit{2)} properly assess the performance. 
Towards these challenges, we propose GCA evaluation for actor LLMs' role-playing abilities, comprising two stages: 
multi-agent simulation and penalty-based LLM judging, as illustrated in Fig.~\ref{fig:main}.

\textbf{Multi-agent Simulation} \quad 
For a test conversation $M$, 
we build a multi-agent system to simulate a conversation $\bar{M}$, in the same setting as $M$. 
We create an RPLA $\agent_\persona$ for each character $c \in \mathcal{C}$ using the actor LLM.
We provide RPLAs with comprehensive data as described in \S\ref{sec:training}:   
scenario descriptions and involved character profiles offer crucial context, and character motivations promote RPLA proactiveness and a natural conversation flow. 
Following \S\ref{sec:data_design}, 
RPLAs are instructed to output in the speech-action-thought format. 
Each RPLA's motivations and inner thoughts are inaccessible to other RPLAs. 
We adopt an NSP model to select the speaker of each turn from $\mathcal{C}\cup\{e\}$, and another LLM as the environment model $\agent_e$ to provide environmental feedback. 
The simulation ends upon an $<$END$>$ signal from NSP, or reaching the maximum of 20 turns. 
In this way, we obtain a multi-turn, multi-character simulation that comprehensively reflects the actor LLMs' role-playing abilities. 

In addition, we introduce a continue-from parameter $k$, where the simulation starts from the first $k$ original messages in $\mathcal{M}$. 
Setting $k>0$ controls the story direction and language style, similar to in-context learning. 
Hence, it enables more controlled evaluation and reduces the influence of different language styles of LLMs. 

\textbf{Penalty-based LLM Judging} \quad 
In this stage, we assess the simulated dialogue $\bar{M}$ via LLM critics. 
Different from previous LLM-as-a-judge methods for RPLA evaluation, our LLM critics:  
\textit{1)} apply penalty-based scoring by identifying role-playing flaws following detailed rubrics, and 
\textit{2)} leverage the original conversation $M$ as reference. 

Specifically, we employ LLM critics~\footnote{This paper uses GPT-4o as the critic LLM by default. } to identify  flaw instances $\mathcal{F}$ in $\bar{M}$ of specific rubrics, such as  \dq{\textit{deviate from the original conversation}} or \dq{\textit{lack initiative and goals}}, instead of directly outputting a score in previous work~\citep{wang2023rolellm, tu2024charactereval}. 
Each flaw $f$ is assigned a severity $v_f$ from 1 (minor) to 5 (severe). 
The initial score for each dimension is calculated as $s = 100-5*\sum_{f\in\mathcal{F}}v_f$. 

\begin{figure}[h]
    \centering
    \includegraphics[width=0.5\linewidth]{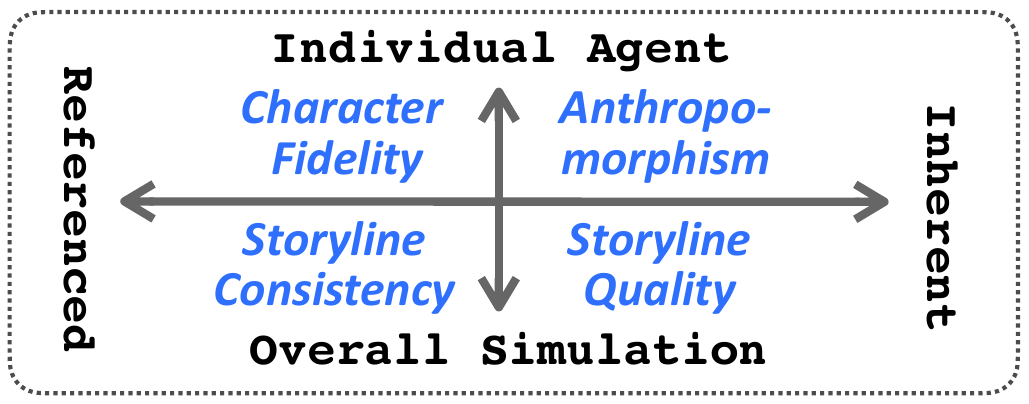}
    \caption{
    We divide the RPLA evaluation into four quadrants (dimensions).
The X-axis represents the evaluation perspective: Referenced (comparing with data in the source book) versus Inherent (assessing standalone quality). 
The Y-axis indicates the evaluation scope: Individual Agent versus Overall Simulation. 
    }
    \label{fig:dims}
\end{figure}

We consider four evaluation dimensions, as shown in Fig. ~\ref{fig:dims}. 
The rubrics are derived from human-annotated issues in extensive human-RPLA conversations from real users and our multi-agent simulations, as well as dimensions from previous work~\citep{shanahan2023role, shao2023character, chen2024from}. 
For more informed evaluation, LLM critics are provided with additional materials, \ie, the original conversation $M$ and plot summary, besides data available to actor LLMs. 
Each dimension is assessed in independent LLM requests. 
The evaluation dimensions and their rubrics are summarized as follows:
\begin{enumerate}[itemsep=-3pt, topsep=0pt, partopsep=0pt]
    \item \textbf{Anthropomorphism}: Evaluates whether RPLAs behave in a human-like manner, with rubrics covering self-identity, emotional depth, persona coherence, and social interaction. 
    \item \textbf{Character Fidelity}: Assesses whether RPLAs faithfully portray their characters, with rubrics examining language style, knowledge and background, personality and behavior, and social relationships. 
    \item \textbf{Storyline Quality}: Evaluates whether the simulated conversation develops naturally, with rubrics focusing on narrative flow and logical consistency. 
    \item \textbf{Storyline Consistency}:  Measures alignment between the simulated conversation $\bar{M}$ and original dialogue $M$, \ie, whether RPLAs' reactions (emotions, attitudes, behaviors) remain consistent with the original. 
\end{enumerate} 

As longer simulations naturally make more flaws, we implement length correction to reduce bias in LLM judging following ~\citet{dubois2024alpacaeval}. 
Specifically, we obtain the length corrected score as 
$s = 100-5*\sum_{f\in\mathcal{F}}v_f + \lambda|\bar{M}|$, 
where $\lambda$ is set to 1.5 based on statistical analysis in \S\ref{sec:length_correction}. 
For detailed prompts and rubrics, please refer to \S\ref{sec:prompts}.

\section{Experiments}

\subsection{Settings}

\textbf{Evaluation Protocol} \quad 
We evaluate LLMs' role-playing abilities through GCA on \method Test, a test set of held-out conversations from the final 10\% of each book. 
\method Test contains 200 conversations, with 100 from books used in \method training and 100 otherwise.  
We employ GPT-4o as the critic LLM and environment model, and \method 70B for NSP. 
We exclude characters' inner thoughts for LLM critics, and set the continue-from parameter $k=0$. 
The details are in \S\ref{sec:exp_setting}.

\textbf{Metrics} \quad 
We report LLM-judged scores for each dimension, and their average as the overall score.
For analysis, we also evaluate two traditional metrics based on N-gram, \ie,  BLEU~\citep{papineni2002bleu} and ROUGE-L\citep{lin2004rougel} compared against original dialogues. 
Additionally, we report win rates versus GPT-3.5 and GPT-4o in \S\ref{sec:additional_results}. %

\textbf{Models} \quad 
Our experiments cover numerous LLMs: 
\textit{1)} \textbf{Close models}, including 
Minimax Abab7-preview~\footnote{https://www.minimaxi.com/}, Doubao-pro~\footnote{https://team.doubao.com/en/ Version:241215}, Step-1-Flash and Step-2~\footnote{https://www.stepfun.com/ Version: 241111 for Step-1-Flash, 241223 for Step-2 (internal).}, GPT-3.5~\citep{openai2022chatgpt}, GPT-4o and GPT-4o Mini~\footnote{https://openai.com/index/hello-gpt-4o/ Version:240806}, Gemini-1.5-Pro~\footnote{https://deepmind.google/technologies/gemini/pro/}, Claude-3-Haiku and Claude-3.5-Sonnet~\footnote{https://www.anthropic.com/ Version: 20240307 for Claude-3-Haiku, 20240620 for Claude-3.5-Sonnet}; 
\textit{2)} \textbf{Open models}, including \method 8B and 70B, LLaMA-3.1-Instruct 8B and 70B~\citep{dubey2024llama},  Qwen-2-Instruct 7B and  72B~\citep{yang2024qwen2technicalreport}, Vicuna-13B-1.5~\citep{zheng2023judging}, Mixtral-8x7B~\citep{jiang2024mixtral}, DeepSeek-V3~\citep{liu2024deepseek} and Higgs-Llama-3-70B~\footnote{https://boson.ai/higgs-v2/}.

\textbf{Other Benchmarks} \quad 
We also evaluate \method models on existing RPLA benchmarks based on multi-choice questions instead of LLM judges, including InCharacter~\citep{wang2024incharacter} for personality tests, LifeChoice~\citep{xu2024character} for decision-making, and CroSS~\citep{yuan2024evaluating} for motivation recognition. 
For InCharacter, we report its accuracy on the Big Five Inventory (BFI).  

\subsection{Main Results}

\begin{table*}[t]
\centering
\setlength{\tabcolsep}{2.5pt}
\renewcommand{\arraystretch}{0.9}
{\small
\begin{tabularx}{\textwidth}{@{}l*{4}{c}ccc@{}}
\toprule
\multirow{3}{*}{\textbf{Model}} & \multicolumn{5}{c}{\textbf{Based on LLM Judges}} & \multicolumn{2}{c}{\textbf{Based on N-gram}} \\ \cmidrule(lr){2-6} \cmidrule(lr){7-8} & \multirow{2}{*}{\makecell{\textbf{Storyline}\\\textbf{Consistency}}} & \multirow{2}{*}{\makecell{\textbf{Anthro-}\\\textbf{pomorphism}}} & \multirow{2}{*}{\makecell{\textbf{Character}\\\textbf{Fidelity}}} & \multirow{2}{*}{\makecell{\textbf{Storyline}\\\textbf{Quality}}} & \multirow{2}{*}{\makecell{\textbf{Average}\\\textbf{Score}}} & \multirow{2}{*}{\makecell{\textbf{BLEU}}} & \multirow{2}{*}{\makecell{\textbf{ROUGE-L}}} \\
 &  &  &  &  &  &  &  \\
\midrule
\multicolumn{8}{c}{\textit{Close-source Models}} \\
\midrule
Abab7-preview & 56.81\scriptsize{$\pm1.47$} & 44.23\scriptsize{$\pm1.90$} & 43.83\scriptsize{$\pm2.71$} & 74.83\scriptsize{$\pm0.97$} & 54.92\scriptsize{$\pm0.57$} & 4.96\scriptsize{$\pm0.07$} & 11.50\scriptsize{$\pm0.06$} \\
Doubao-pro & 60.95\scriptsize{$\pm1.40$} & 49.72\scriptsize{$\pm0.23$} & 47.02\scriptsize{$\pm1.10$} & 79.28\scriptsize{$\pm0.82$} & 59.24\scriptsize{$\pm0.30$} & 6.38\scriptsize{$\pm0.08$} & 12.95\scriptsize{$\pm0.04$} \\
Step-1-Flash & 57.75\scriptsize{$\pm0.72$} & 48.12\scriptsize{$\pm0.39$} & 44.48\scriptsize{$\pm0.48$} & 75.93\scriptsize{$\pm0.99$} & 56.57\scriptsize{$\pm0.48$} & 5.95\scriptsize{$\pm0.15$} & 12.71\scriptsize{$\pm0.11$} \\
Step-2 & 61.43\scriptsize{$\pm0.88$} & 49.06\scriptsize{$\pm1.69$} & 47.33\scriptsize{$\pm0.70$} & 77.96\scriptsize{$\pm0.85$} & 58.94\scriptsize{$\pm0.75$} & 5.75\scriptsize{$\pm0.08$} & 12.50\scriptsize{$\pm0.11$} \\
GPT-3.5 & 57.22\scriptsize{$\pm0.13$} & 43.30\scriptsize{$\pm0.48$} & 42.29\scriptsize{$\pm1.47$} & 73.91\scriptsize{$\pm0.64$} & 54.18\scriptsize{$\pm0.63$} & 4.58\scriptsize{$\pm0.11$} & 11.80\scriptsize{$\pm0.10$} \\
GPT-4o & \textbf{61.59\scriptsize{$\pm0.66$}} & 48.93\scriptsize{$\pm0.48$} & \textbf{48.95\scriptsize{$\pm1.73$}} & \textbf{80.33\scriptsize{$\pm0.59$}} & \textbf{59.95\scriptsize{$\pm0.50$}} & 5.90\scriptsize{$\pm0.16$} & 12.11\scriptsize{$\pm0.13$} \\
GPT-4o Mini & 60.09\scriptsize{$\pm0.60$} & 48.21\scriptsize{$\pm1.09$} & 44.88\scriptsize{$\pm1.63$} & 78.55\scriptsize{$\pm0.14$} & 57.93\scriptsize{$\pm0.74$} & 3.90\scriptsize{$\pm0.07$} & 10.81\scriptsize{$\pm0.07$} \\
Gemini Pro & 59.11\scriptsize{$\pm0.82$} & 52.41\scriptsize{$\pm0.57$} & 47.83\scriptsize{$\pm0.37$} & 77.59\scriptsize{$\pm1.43$} & 59.24\scriptsize{$\pm0.25$} & 5.39\scriptsize{$\pm0.04$} & 11.65\scriptsize{$\pm0.06$} \\
Claude-3-Haiku & 58.18\scriptsize{$\pm0.72$} & 44.66\scriptsize{$\pm1.72$} & 41.88\scriptsize{$\pm0.34$} & 74.14\scriptsize{$\pm1.26$} & 54.71\scriptsize{$\pm0.84$} & 4.80\scriptsize{$\pm0.05$} & 12.02\scriptsize{$\pm0.02$} \\
Claude-3.5-Sonnet & 57.45\scriptsize{$\pm0.98$} & 48.50\scriptsize{$\pm2.35$} & 45.69\scriptsize{$\pm1.80$} & 77.23\scriptsize{$\pm0.88$} & 57.22\scriptsize{$\pm0.95$} & 5.17\scriptsize{$\pm0.12$} & 11.45\scriptsize{$\pm0.07$} \\
\midrule
\multicolumn{8}{c}{\textit{Open-source Models}} \\
\midrule
Mistral-7B & \underline{59.90\scriptsize{$\pm1.33$}} & 40.00\scriptsize{$\pm0.74$} & 44.75\scriptsize{$\pm1.14$} & 61.93\scriptsize{$\pm1.12$} & 51.64\scriptsize{$\pm0.55$} & 2.71\scriptsize{$\pm0.10$} & 9.28\scriptsize{$\pm0.12$} \\
Qwen-2-7B & 51.96\scriptsize{$\pm0.67$} & 35.48\scriptsize{$\pm0.62$} & 31.51\scriptsize{$\pm2.95$} & 63.18\scriptsize{$\pm0.79$} & 45.53\scriptsize{$\pm0.69$} & 4.21\scriptsize{$\pm0.21$} & 10.71\scriptsize{$\pm0.10$} \\
LLaMA-3.1-8B & 54.10\scriptsize{$\pm1.63$} & 45.36\scriptsize{$\pm1.91$} & 40.22\scriptsize{$\pm1.16$} & 72.29\scriptsize{$\pm1.75$} & 52.99\scriptsize{$\pm1.20$} & 4.59\scriptsize{$\pm0.11$} & 10.18\scriptsize{$\pm0.09$} \\
CoSER-8B & 58.61\scriptsize{$\pm2.46$} & 47.23\scriptsize{$\pm0.16$} & 46.90\scriptsize{$\pm2.06$} & 73.04\scriptsize{$\pm1.37$} & 56.45\scriptsize{$\pm0.56$} & 9.40\scriptsize{$\pm0.18$} & 14.21\scriptsize{$\pm0.11$} \\
Vicuna-13B-1.5 & 52.75\scriptsize{$\pm1.64$} & 39.12\scriptsize{$\pm1.21$} & 38.04\scriptsize{$\pm0.98$} & 60.43\scriptsize{$\pm1.58$} & 47.58\scriptsize{$\pm1.25$} & 1.67\scriptsize{$\pm0.10$} & 5.59\scriptsize{$\pm0.18$} \\
Mixtral-8x7B & 51.25\scriptsize{$\pm1.73$} & 38.44\scriptsize{$\pm1.18$} & 36.92\scriptsize{$\pm2.65$} & 67.69\scriptsize{$\pm0.80$} & 48.58\scriptsize{$\pm1.35$} & 5.28\scriptsize{$\pm0.06$} & 11.66\scriptsize{$\pm0.05$} \\
Qwen-2-72B & 57.75\scriptsize{$\pm1.26$} & 47.28\scriptsize{$\pm0.87$} & 46.62\scriptsize{$\pm1.69$} & 76.60\scriptsize{$\pm0.36$} & 57.06\scriptsize{$\pm1.00$} & 5.38\scriptsize{$\pm0.00$} & 11.85\scriptsize{$\pm0.03$} \\
LLaMA-3.1-70B & 57.46\scriptsize{$\pm1.65$} & 45.95\scriptsize{$\pm1.30$} & 43.72\scriptsize{$\pm1.17$} & 74.84\scriptsize{$\pm0.54$} & 55.49\scriptsize{$\pm0.33$} & 4.82\scriptsize{$\pm0.06$} & 10.98\scriptsize{$\pm0.06$} \\
Higgs-Llama-3-70B & 57.10\scriptsize{$\pm1.12$} & 43.82\scriptsize{$\pm2.18$} & 42.41\scriptsize{$\pm1.66$} & 75.62\scriptsize{$\pm0.15$} & 54.74\scriptsize{$\pm1.26$} & 3.99\scriptsize{$\pm0.33$} & 10.92\scriptsize{$\pm0.56$} \\
CoSER-70B & 58.66\scriptsize{$\pm1.34$} & \underline{\textbf{53.33\scriptsize{$\pm0.91$}}} & \underline{48.75\scriptsize{$\pm1.43$}} & 75.49\scriptsize{$\pm0.94$} & \underline{59.06\scriptsize{$\pm0.22$}} & \underline{\textbf{10.10\scriptsize{$\pm0.04$}}} & \underline{\textbf{14.78\scriptsize{$\pm0.09$}}} \\
DeepSeek-V3 & 56.40\scriptsize{$\pm0.95$} & 47.87\scriptsize{$\pm1.10$} & 44.02\scriptsize{$\pm0.13$} & \underline{76.66\scriptsize{$\pm1.26$}} & 56.24\scriptsize{$\pm0.46$} & 4.54\scriptsize{$\pm0.14$} & 11.02\scriptsize{$\pm0.15$} \\
\bottomrule
\end{tabularx}
}
\caption{Performance (\%) of various LLMs on given-circumstance acting using \method Test. 
\textbf{Bold} or \underline{underlined} values indicate best  performance across all models and open-source models, respectively. 
}
\label{tab:model-comparison}
\end{table*}

\textbf{Performance of Various LLMs on \method Test} \quad 
We apply \method Test to evaluate extensive LLMs. 
The results shown in Table~\ref{tab:model-comparison} are  averaged across three runs, from which we observe that: 
\textit{1)} \method 70B achieves state-of-the-art performance across both LLM-judged  and N-gram-based metrics. 
For LLM-judged metrics, \method 70B outperforms all open models and shows competitive performance with GPT-4o. \method 8B similarly outperforms models of comparable scale.
For N-gram-based metrics, \method models demonstrate substantial improvements over existing models, exceeding the second-best performance by 58\% on BLEU; 
\textit{2)} Among all models, GPT-4o, Gemini Pro, Claude-3.5-Sonnet, Doubao-pro, Step-2, Qwen-2-72B, and \method 70B demonstrate superior performance, achieving average scores above 57\%; 
\textit{3)} Table~\ref{tab:exp_idood} presents LLM performance separately for test splits from books included or excluded for  \method training. 
The results show consistent trends across both splits, validating that \method models maintain strong performance on out-of-domain characters. 

\textbf{Authentic Conversations from High-quality Novels Improve LLMs' Role-playing Ability} \quad 
According to Table \ref{tab:model-comparison}, 
\method models demonstrate significant improvements over their LLaMA 3.1 baselines.
In contrast, Higgs-LLaMA-3 70B, fine-tuned on synthesized dialogues, performs below LLaMA-3.1 70B,
These results highlight the importance of high-quality, authentic role-playing data for LLM training. 

\textbf{Conversation Continuation Enables More Controlled Evaluation} \quad 
Table ~\ref{tab:model-comparison-cf3} shows experiment results when multi-agent systems start from the first $k=3$ original messages.  
In this setting, model obtain higher scores compared to simulations from scratch ($k=0$), with reduced performance gaps between different models, especially for BLEU and ROUGE-L results. 
For example, the average score gap between Qwen-2 72B and 7B decreases from 11.5\% ($k=0$) to 8.8\% ($k=3$).
This improvement occurs because the $k=3$ original messages guide the story direction and language style, particularly benefiting smaller models that typically struggle with complex role-playing instructions.

\textbf{Results on Other Benchmarks} \quad 
We evaluate \method and other models on existing benchmarks for RPLAs based on multi-choice questions. 
As shown in Table~\ref{tab:exp_other_bmks}, \method 70B achieves state-of-the-art performance across these benchmarks. 
Notably, \method 70B achieves 93.47\% accuracy on LifeChoice, surpassing GPT-4o by 23\%. 
These results exhibit \method models' strong capability in nuanced portrayal of characters' personalities and behaviors. 

\begin{table}[t]
\centering
\setlength{\tabcolsep}{4pt}
\renewcommand{\arraystretch}{0.9}
\begin{tabular}{lcccc}
\toprule
\multirow{2}{*}{\textbf{Model}} & \multicolumn{2}{c}{\textbf{Incharacter}} & \multirow{2}{*}{\makecell{\textbf{Life}\\\textbf{Choice}}} & \multirow{2}{*}{\makecell{\textbf{CroSS}\\\textbf{MR}}} \\
 & \textbf{Dim} & \textbf{Full} &  &  \\
\midrule
LLaMA-3.1-8B & 64.97 & 15.62 & 61.10 & 30.15 \\
CoSER-8B & 75.80 & 21.88 & 69.54 & 44.94 \\
 \hfill\hfill\hfill {\textit{trained w/o} \textsc{I.T.}} & 70.70 & 15.62 & 59.92 & 43.14 \\
LLaMA-3.1-70B & 72.16 & 31.25 & 86.48 & 61.30 \\
Higgs-Llama-3-70B & 74.52 & 28.12 & 74.03 & 60.12 \\
CoSER-70B & 75.80 & \textbf{34.38} & \textbf{93.47} & \textbf{64.49} \\
 \hfill\hfill\hfill {\textit{trained w/o} \textsc{I.T.}} & 73.12 & 32.14 & 93.18 & 63.14 \\
Qwen-2-72B & 74.52 & 31.25 & 81.14 & 62.57 \\
GPT-3.5 & 71.20 & 21.88 & 78.07 & 30.09 \\
GPT-4o & \textbf{76.54} & 32.62 & 75.96 & \textbf{64.49} \\
Claude-3.5-Sonnet & 72.61 & 21.88 & 86.07 & 30.59 \\
\bottomrule
\end{tabular}
\caption{LLM performance (\%) across three existing RPLA benchmarks. I.T. denotes inner thoughts. For InCharacter, we report accuracy for individual (Dim) and full (Full) dimensions on BFI. }
\label{tab:exp_other_bmks}
\end{table}

\subsection{Human Evaluation}
\label{sec:human_eval}

We conduct human evaluation
to further validate the performance of \method models and the reliability of LLM critics in GCA evaluation. 

\paragraph{Annotations}
We randomly select 60 samples from \method Test, and evaluate 7 representative LLMs (shown in Table~\ref{tab:human_eval}) based on their simulation results. 
Three annotators independently score the LLMs on 20 samples each, using a 1 to 10 scoring scale,  given the background context and original conversations. During this process, the annotators note that 
\textit{1)} GCA simulation deeply reflects LLMs' role-playing abilities, compared with previous single-turn benchmarks; \textit{2)} Manual evaluation is highly challenging and time-consuming. It requires careful learning of complex background and abundant dialogues, and takes 15 minutes to evaluate 7 models for one sample on average. 
We report models' average scores and win rate against other models. 

\begin{table}[htbp]
\centering
\renewcommand{\arraystretch}{0.9}
\begin{tabular}{lcc}
\toprule
\textbf{Model} & \textbf{Score} & \textbf{Win Rate (\%)} \\
\midrule
GPT-3.5 & 3.117 & 10.6 \\
LLaMA-3.1-8B & 3.600 & 19.4 \\
Abab7-preview & 4.533 & 37.5 \\
\method-8B & 4.567 & 38.6 \\
GPT-4o & 4.967 & 47.2 \\
Claude-3.5-Sonnet & 6.200 & 73.9 \\
\method-70B & \textbf{6.783} & \textbf{86.9} \\
\bottomrule
\end{tabular}
\caption{Human evaluation on LLMs' GCA simulation.}
\label{tab:human_eval}
\end{table}

\paragraph{Results and Analysis} 
According to  Table~\ref{tab:human_eval}, we find that: 
\textit{1)} The results generally align with GCA evaluation, validating the reliability of our LLM-based evaluation protocol;
\textit{2)} \method 70B maintain  superior performance, obtaining the highest score (6.783) and win rate (86.9\%);
\textit{3)} Human evaluators show less preference for GPT models compared to GPT-4o as LLM judges, which likely stems from self-preference bias in LLM judges~\citep{wataoka2024self}.

\paragraph{Alignment} 
We further study the alignment between LLM and human judges.
We evaluate three judge models, GPT-4o, DeepSeek-V3 and DeepSeek-R1~\citep{guo2025deepseek}.
We consider standard GCA evaluation and ablation variants that remove: \textit{(i)} original dialogues as reference, \textit{(ii)} scoring rubrics, \textit{(iii)} length correction, and \textit{(iv)} dimension separation.
We also compare with BLEU and ROUGE-L.
The alignment is measured via model-wise preferences (\S\ref{sec:appendix_human_eval}). 

The results in Table~\ref{tab:judge_alignment} reveal that:
\textit{1)} All components of GCA evaluation contribute to improved alignment;
\textit{2)} Reasoning models excel as judges, with DeepSeek-R1 achieving the highest alignment;
\textit{3)} BLEU and ROUGE-L remain effective indicators for role-playing evaluation.

\begin{table}[htbp]
  \centering
  \renewcommand{\arraystretch}{0.9}
  \begin{tabular}{lc}
  \toprule
  \textbf{Scoring Method} & \textbf{Alignment} \\
  \midrule
  GCA (GPT-4o) & 68.6 \\
  \quad\quad {\textit{w/o} reference} & 64.3 \\
  \quad\quad {\textit{w/o} rubrics} & 65.1 \\
  \quad\quad {\textit{w/o} length correction} & 64.5 \\
  \quad\quad {\textit{w/o} dimension separation} & 65.2 \\
  GCA (DeepSeek-V3) & 65.1 \\
  GCA (DeepSeek-R1) & \textbf{77.5} \\
  \quad\quad {\textit{w/o} reference}  & 77.2 \\
  \midrule
  BLEU  & 75.3 \\
  ROUGE-L  & 72.0 \\
  \bottomrule
  \end{tabular}
  \caption{Alignment (\%) between human judges and LLM judges (or automatic metrics).}
  \label{tab:judge_alignment}
  \end{table}

\subsection{Ablation Studies}

\begin{table}[t]
\centering
\setlength{\tabcolsep}{3pt}
\renewcommand{\arraystretch}{0.9}
\begin{tabular}{lccc}%
\toprule
\multirow{1}{*}{\textbf{Model}} & \multicolumn{1}{c}{\textbf{Standard}} & \multicolumn{1}{c}{\textbf{Test w/o I.T.}} & \multicolumn{1}{c}{\textbf{Test w/o Mot.}} \\
\midrule
GPT-4o & 59.95 & 56.89 & 56.34 \\
Qwen-2-72B & 57.06 & 51.95 & 54.21 \\
LLaMA-3.1-70B & 55.49 & 53.12 & 52.49 \\
CoSER-70B & 59.06 & 57.32 & 57.71 \\
\hfill\hfill\hfill {\textit{trained w/o} \textsc{I.T.}} & 56.04 & 55.34 & - \\
LLaMA-3.1-8B & 52.99 & 51.97 & 49.63 \\
CoSER-8B & 56.45 & 54.65 & 56.81 \\
\hfill\hfill\hfill {\textit{trained w/o} \textsc{I.T.}} & 54.25 & 54.38 & - \\
\bottomrule
\end{tabular}
\caption{
Ablation study results (average scores) on \method Test. 
I.T. and Mot. mean inner thoughts and motivations, respectively. }
\label{tab:model-comparison-inner-thought}
\end{table}

\textbf{Inner Thoughts and Motivations Enhance RPLAs at Test Time} \quad 
Table~\ref{tab:model-comparison-inner-thought} compares LLMs' overall scores on \method Test with or without inner thoughts and  motivations. 
The results show consistent performance improvements across all models when inner thoughts and motivations are included.

\textbf{Inner Thoughts Benefit Role-Playing Training} \quad 
We train \method model variants without inner thoughts, and evaluate them on various benchmarks.  
Results in Tables ~\ref{tab:exp_other_bmks} and ~\ref{tab:model-comparison-inner-thought}  show that models trained without inner thoughts consistently underperform regular \method models, exhibiting the value of inner thoughts for role-playing training.

\subsection{\method Dataset for Retrieval Augmentation}

We evaluate the value of our comprehensive data types for retrieval augmentation on \method Test. 
We explore three retrieval sources for a specific character: 
dialogues in related conversations (Conv.), as well as experiences (Expr.) and raw text in related plots. 
The retrieval system is based on FAISS~\citep{douze2024faiss} with BGE-M3~\citep{chen2024bge} for text embeddings.
As shown in Fig.~\ref{tab:rag}, we observe:
\textit{1)} Models consistently benefit from characters' retrieved experiences and conversations, especially for \method 70B;
\textit{2)} However, raw text retrieval barely enhances LLMs' performance. 
Detailed experimental settings and results are provided in \S\ref{sec:additional_results} and Table ~\ref{tab:rag}.

\begin{figure}[t]
    \centering
    \includegraphics[width=0.7\textwidth]{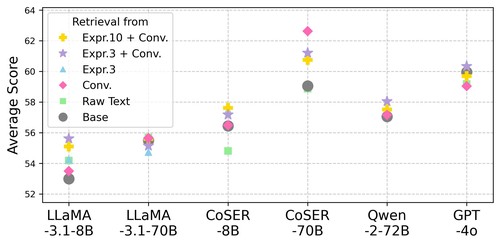}
    \caption{LLM Performance on \method Test 
    with retrieval augmentation from various character data. 
    Expr. and Conv. denote experiences and conversations respectively.  
    }
    \label{fig:distribution}
\end{figure}

\subsection{Case Studies}

We conduct case studies to analyze LLMs' performance in GCA simulation. 
Several cases are presented in Tables~\ref{tab:case_sansa} to~\ref{tab:case_cerci4}, from which we observe that:
\textit{1)} \method models, trained on authentic dialogues, communicate more naturally, closely aligning with human speech patterns.
\textit{2)} \method models better recall character-related knowledge, such as the iconic line \textit{\dq{Grown enough to be wed, wed enough to be bedded}} by \textit{Lysa Arryn} in Table~\ref{tab:case_sansa}.
This is consistent with their high BLEU and ROUGE-L scores.
\textit{3)} \method models better portray sophisticated thinking process of humans. 
For example, in Tables~\ref{tab:case_cerci} to~\ref{tab:case_cerci4}, 
\method 70B faithfully reproduces \textit{Cersei Lannister}’s suppressed anger as depicted in the original conversations, while other models, including GPT-4o and Claude-3.5-Sonnet, resort to a stereotypical portrayal of her arrogance and pride.

\section{Conclusion}
Towards effective RPLAs for established characters, 
this paper introduces \method, 
a collection of an authentic dataset, along with models and evaluation protocol based on such data.
The \method dataset offers high-quality data from 771 renowned books, and includes comprehensive data types such as authentic dialogues, plot summaries, character experiences, inner thoughts, \etc.
Then, we propose given-circumstance acting (GCA) for training and evaluating role-playing LLMs, where LLMs sequentially portray multiple characters within authentic book scenarios. 
Applying GCA training to LLaMA-3.1 models using our dataset, we develop \method 8B and \method 70B, advanced open LLMs for role-playing. 
For evaluation, 
GCA combines multi-agent simulation and penalty-based LLM critics. 
Extensive experiments exhibit \method dataset's value for RPLA training, evaluation, and retrieval.
Moreover, \method models achieve state-of-the-art performance on both our evaluation and three existing RPLA benchmarks.

\section*{Impact Statement} %

\method aims to advance RPLA research by providing effective dataset, models and evaluation protocol. 
We will release our dataset, models, and evaluation scripts to foster innovation in RPLAs. 
The dataset is intended for research purposes only.
For copyright policies, we will not distribute raw novel content. 
We require that anyone using our work must adhere to copyright policies and obtain proper permissions for any derivative works. 
The \method dataset is derived from literary works may involve ethical considerations, and the content involved does not represent the authors' viewpoints. 
Our methods can potentially be applied to develop agents for real-world individuals. 
However, such applications must strictly respect personal data privacy and obtain necessary consent.
We hope our research will benefit RPLA researchers and developers, but emphasize the importance of responsible development. Any applications must respect copyright policies, personal data privacy, and be developed with proper licensing.

\bibliography{example_paper}

\newpage
\appendix
\onecolumn
\onecolumn
\appendix

\section{Dataset}
\label{sec:app_dataset}

\subsection{Statistics and Analysis}
\label{sec:data_statistics}
As shown in Table \ref{tab:dataset_overview}, \method dataset is extensive and comprehensive, encompassing dialogue data from 771 books and 17,966 distinct characters. 
The dataset includes 30,069 unique plots and 29,798 conversations. 
On average, each conversation consists of approximately 13.2 utterances, with the entire dataset comprising a total of 392,298 utterances.

\begin{table}[htbp]
\centering
\begin{tabular}{|c|c|c|c|c|}
\hline
\textbf{\#Book} & \textbf{\#Plot} & \textbf{\#Conversation.} & \textbf{\#Chararacter} & \textbf{\#Utterance} \\ \hline
771     &   30,069      & 29,798                   & 17,966               & 392,298              \\ \hline
\end{tabular}
\caption{Statistics of \method Dataset.
}
\label{tab:dataset_overview}
\end{table}

Our book selection is derived from the \textit{Best Books Ever} list on \textit{Goodreads}, a curated collection of globally acclaimed literary works. 
These novels have garnered widespread recognition and appreciation from readers worldwide.
Table ~\ref{tab:selected_books} presents a comprehensive list of the top 100 books from our selection.

\begin{table*}[htbp]
\small
\centering
\resizebox{\linewidth}{!}{
\begin{tabular}{p{3.2in}|p{3.2in}}
\toprule
\multicolumn{2}{c}{\textbf{Selected Books}} \\  \hline
\textbf{1}. \textit{{The Hunger Games (The Hunger Games, \#1)}} & \textbf{2}. \textit{{Harry Potter and the Order of the Phoenix (H. P., \#5)}}\\ \hline
\textbf{3}. \textit{{Pride and Prejudice}} & \textbf{4}. \textit{{To Kill a Mockingbird}}\\ \hline
\textbf{5}. \textit{{The Book Thief}} & \textbf{6}. \textit{{Animal Farm}}\\ \hline
\textbf{7}. \textit{{The Chronicles of Narnia (\#1-7)}} & \textbf{8}. \textit{{The Fault in Our Stars}}\\ \hline
\textbf{9}. \textit{{The Picture of Dorian Gray}} & \textbf{10}. \textit{{Wuthering Heights}}\\ \hline
\textbf{11}. \textit{{Gone with the Wind}} & \textbf{12}. \textit{{The Perks of Being a Wallflower}}\\ \hline
\textbf{13}. \textit{{The Lightning Thief (Percy Jackson and the Olympians, \#1)}} & \textbf{14}. \textit{{The Little Prince}}\\ \hline
\textbf{15}. \textit{{The Great Gatsby}} & \textbf{16}. \textit{{Crime and Punishment}}\\ \hline
\textbf{17}. \textit{{Memoirs of a Geisha}} & \textbf{18}. \textit{{Les Misérables}}\\ \hline
\textbf{19}. \textit{{The Alchemist}} & \textbf{20}. \textit{{Lord of the Flies}}\\ \hline
\textbf{21}. \textit{{The Hitchhiker’s Guide to the Galaxy (\#1)}} & \textbf{22}. \textit{{The Help}}\\ \hline
\textbf{23}. \textit{{Dracula}} & \textbf{24}. \textit{{Ender’s Game (Ender's Saga, \#1)}}\\ \hline
\textbf{25}. \textit{{Of Mice and Men}} & \textbf{26}. \textit{{One Hundred Years of Solitude}}\\ \hline
\textbf{27}. \textit{{Brave New World}} & \textbf{28}. \textit{{A Thousand Splendid Suns}}\\ \hline
\textbf{29}. \textit{{The Time Traveler’s Wife}} & \textbf{30}. \textit{{The Princess Bride}}\\ \hline
\textbf{31}. \textit{{The Secret Garden}} & \textbf{32}. \textit{{The Outsiders}}\\ \hline
\textbf{33}. \textit{{A Game of Thrones (A Song of Ice and Fire, \#1)}} & \textbf{34}. \textit{{Little Women}}\\ \hline
\textbf{35}. \textit{{A Wrinkle in Time (Time Quintet, \#1)}} & \textbf{36}. \textit{{The Odyssey}}\\ \hline
\textbf{37}. \textit{{Harry Potter and the Deathly Hallows (H. P., \#7)}} & \textbf{38}. \textit{{Frankenstein: The 1818 Text}}\\ \hline
\textbf{39}. \textit{{The Kite Runner}} & \textbf{40}. \textit{{The Handmaid’s Tale (The Handmaid's Tale, \#1)}}\\ \hline
\textbf{41}. \textit{{The Lovely Bones}} & \textbf{42}. \textit{{The Adventures of Huckleberry Finn}}\\ \hline
\textbf{43}. \textit{{Life of Pi}} & \textbf{44}. \textit{{A Tale of Two Cities}}\\ \hline
\textbf{45}. \textit{{Dune (Dune, \#1)}} & \textbf{46}. \textit{{Harry Potter and the Prisoner of Azkaban (H.P.,\#3)}}\\ \hline
\textbf{47}. \textit{{Water for Elephants}} & \textbf{48}. \textit{{Harry Potter and the Sorcerer’s Stone (H. P., \#1)}}\\ \hline
\textbf{49}. \textit{{The Bell Jar}} & \textbf{50}. \textit{{Matilda}}\\ \hline
\textbf{51}. \textit{{The Stand}} & \textbf{52}. \textit{{Catch-22}}\\ \hline
\textbf{53}. \textit{{The Adventures of Sherlock Holmes (S. H., \#3)}} & \textbf{54}. \textit{{The Pillars of the Earth (Kingsbridge, \#1)}}\\ \hline
\textbf{55}. \textit{{Rebecca}} & \textbf{56}. \textit{{Great Expectations}}\\ \hline
\textbf{57}. \textit{{The Girl with the Dragon Tattoo (Millennium, \#1)}} & \textbf{58}. \textit{{The Color Purple}}\\ \hline
\textbf{59}. \textit{{Anna Karenina}} & \textbf{60}. \textit{{My Sister’s Keeper}}\\ \hline
\textbf{61}. \textit{{The Brothers Karamazov}} & \textbf{62}. \textit{{A Clockwork Orange}}\\ \hline
\textbf{63}. \textit{{And Then There Were None}} & \textbf{64}. \textit{{The Road}}\\ \hline
\textbf{65}. \textit{{To Kill a Mockingbird}} & \textbf{66}. \textit{{The Golden Compass (His Dark Materials, \#1)}}\\ \hline
\textbf{67}. \textit{{Vampire Academy (Vampire Academy, \#1)}} & \textbf{68}. \textit{{Siddhartha}}\\ \hline
\textbf{69}. \textit{{The Complete Stories and Poems}} & \textbf{70}. \textit{{Interview with the Vampire (The Vampire Chronicles, \#1)}}\\ \hline
\textbf{71}. \textit{{Don Quixote}} & \textbf{72}. \textit{{The Old Man and the Sea}}\\ \hline
\textbf{73}. \textit{{The Poisonwood Bible}} & \textbf{74}. \textit{{Harry Potter and the Goblet of Fire (H. P., \#4)}}\\ \hline
\textbf{75}. \textit{{Atlas Shrugged}} & \textbf{76}. \textit{{The Notebook (The Notebook, \#1)}}\\ \hline
\textbf{77}. \textit{{Harry Potter and the Half-Blood Prince (H. P., \#6)}} & \textbf{78}. \textit{{Moby-Dick or, The Whale}}\\ \hline
\textbf{79}. \textit{{A Prayer for Owen Meany}} & \textbf{80}. \textit{{Clockwork Angel (The Infernal Devices, \#1)}}\\ \hline
\textbf{81}. \textit{{The Stranger}} & \textbf{82}. \textit{{The Secret Life of Bees}}\\ \hline
\textbf{83}. \textit{{Harry Potter and the Chamber of Secrets (H. P., \#2)}} & \textbf{84}. \textit{{The Red Tent}}\\ \hline
\textbf{85}. \textit{{The Name of the Wind(The Kingkiller Chronicle,\#1)}} & \textbf{86}. \textit{{The Master and Margarita}}\\ \hline
\textbf{87}. \textit{{The Metamorphosis}} & \textbf{88}. \textit{{Eragon (The Inheritance Cycle, \#1)}}\\ \hline
\textbf{89}. \textit{{The Count of Monte Cristo}} & \textbf{90}. \textit{{Looking for Alaska}}\\ \hline
\textbf{91}. \textit{{The Adventures of Tom Sawyer}} & \textbf{92}. \textit{{Charlie and the Chocolate Factory(Charlie Bucket,\#1)}}\\ \hline
\textbf{93}. \textit{{The Last Olympian (Percy Jackson and the Olympians, \#5)}} & \textbf{94}. \textit{{The Curious Incident of the Dog in the Night-Time}}\\ \hline
\textbf{95}. \textit{{The Shadow of the Wind (Cemetery of Forgotten Books, \#1)}} & \textbf{96}. \textit{{The Unbearable Lightness of Being}}\\ \hline
\textbf{97}. \textit{{On the Road}} & \textbf{98}. \textit{{The Name of the Rose}}\\ \hline
\textbf{99}. \textit{{A Story of Yesterday}} & \textbf{100}. \textit{{The Godfather (The Godfather, \#1)}} 
\\ 
\bottomrule

\end{tabular}}

\caption{The top 100 selected books from Goodreads' \textit{Best Books Ever} list. }
\label{tab:selected_books}
\end{table*}

\begin{figure}[htbp]
    \centering
    \includegraphics[width=0.5\textwidth]{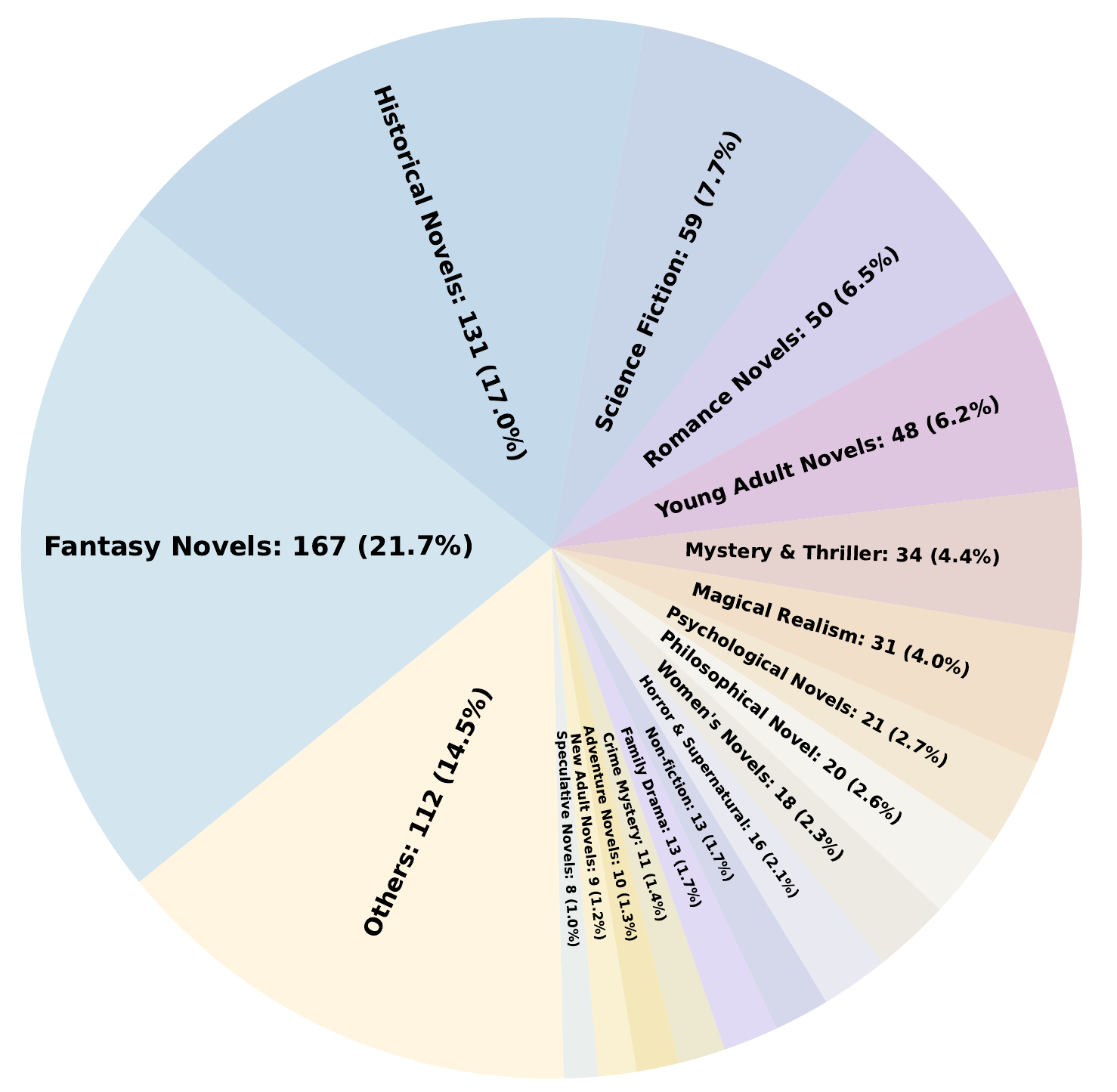}
    \caption{Genre distribution of selected books in \method dataset.
    }
    \label{fig:type}
\end{figure}

We analyze the genres of the selected books based on Supersummary classifications, with the statistical results presented in Figure \ref{fig:type}. 
Our dataset encompasses a wide range of genres, particularly fiction categories such as Fantasy, Historical, Science Fiction, Romance, and Mystery. It also features niche fiction genres, showcasing diverse narrative styles. 
In addition to fiction, the collection includes non-fiction genres such as memoirs, biographies, and other works, enhancing its versatility.

\subsection{Data Splits}
For evaluation purposes, we held out the last 10\% of data from each book; that is, they are not included in our prompts or datasets for training or retrieval purposes. Additionally, we trained the \method models on only 90\% of the books. 

\subsection{Implementation Details for Construction}

\paragraph{Extracting Raw Text} 
LLMs often struggle to extract verbatim original content, especially with punctuation marks like quotation marks, making it difficult to extract raw text directly. 
Therefore, instead of asking LLMs to extract the complete text of a plot, we prompt LLMs to extract the first and last sentences of each plot. 
Since the extracted sentences may still contain typos, we apply lexical similarity to match them with the exact sentences from the raw text. 
Finally, we identify the complete raw text based on these first and last sentences.

\paragraph{Parsing Structured Data} 
During extraction, we instruct LLMs to output extracted data in JSON format. 
However, LLM-generated JSON strings may sometimes be unparseable or do not conform to the specified format (e.g., missing required keys). 

Towards this challenge, we adopt a repair-and-retry strategy to improve extraction success rate.
For each chunk to be extracted, we invoke LLMs and attempt to parse a valid JSON object.
If parsing fails, we employ LLMs to repair the invalid JSON string and retry. 
Some failures occur because the LLM attempts to output JSON that exceeds the maximum token limit, resulting in truncation. 
In such cases, we prompt the repairing LLM to truncate the JSON at an appropriate point. 
If it still fail, we restart the entire process from the beginning, making up to 5 attempts.

\paragraph{Refining Conversation Settings} 
During data extraction, we observe that the initially extracted conversation settings, including scenarios and character motivations, often fail to provide a comprehensive context. 
We attribute this to the LLM’s tendency to distribute information across different data fields when extracting multiple kinds of information simultaneously, rather than repeating it in different data fields. 
For instance, if certain information is already mentioned in the plot summary, it might be omitted from the scenario description.

Therefore, to provide a complete context for given-circumstance acting, we implement an additional LLM call to refine the conversation settings based on the extracted data. 
We instructed the refining LLM to provide a comprehensive  conversation setting, while carefully avoiding any disclosure of subsequent dialogue content or plot developments.

For additional details, such as the regular expressions used for identifying chapter titles, please refer to our  code.

\subsection{Comparison with Existing Methods for Character Profiling}

Previous character profiling methods, including hierarchical updating~\citep{wu2021recursively}, incremental updating~\citep{chang2023booookscore}, and one-shot summarization~\citep{yuan2024evaluating}, typically only generate the profile of a single character at a time. 
Morevoer, ~\citet{papoudakis2024bookworm} shows that these methods, particularly hierarchical updating, perform suboptimally when generating multiple character profiles simultaneously.

\method's multi-stage, extract-then-aggregate pipeline addresses these limitations. 
It ensures comprehensive character profiles with high precision and recall of character knowledge, capturing evolving character arcs, and significantly improving processing efficiency.

\subsection{Safety}
We conduct safety checks on the dataset and remove potentially problematic content. Specifically, we truncate 110 sensitive conversations and remove a total of 602 messages. 

\subsection{Comparison with PersonaHub}

Recently, several studies have explored LLM role-playing using synthesized personas rather, such as PersonaHub~\citep{chan2024personahub}. We provide a systematic comparison between \method and PersonaHub  below.

\paragraph{Goal and Focus}
The fundamental objectives of the two datasets differ significantly:

\begin{table}[h]
\centering
\begin{tabular}{lp{5cm}p{5cm}}
\hline
\textbf{Aspect} & \textbf{\method} & \textbf{PersonaHub} \\
\hline
Target & Simulate established personas with high fidelity & Synthesize instruction data for knowledge distillation \\
\hline
Persona Focus & Depth and richness of character data  & Breadth of persona types \\
\hline
Key Capabilities & Anthropomorphism, character fidelity, multi-character interaction & General instruction following \\
\hline
\end{tabular}
\caption{Comparison between \method and PersonaHub regarding goal and focus.}
\end{table}

\paragraph{Dataset Quality and Design}
The data sources and design reflect distinct research priorities:

\begin{table}[h]
\centering
\begin{tabular}{lp{5cm}p{5cm}}
\hline
\textbf{Aspect} & \textbf{\method} & \textbf{PersonaHub} \\
\hline
Data Source & Authentic book dialogues & GPT-4o synthesized conversations \\
\hline
Persona Origin & Book characters & LLM-synthesized personas \\
\hline
Persona Data & Comprehensive profiles, dialogues,  experiences & Concise profiles with limited context \\
\hline
Design Priority & Quality, authenticity, and character depth & Quantity and diversity\\
\hline
\end{tabular}
\caption{Comparison between \method and PersonaHub regarding dataset quality and design.}
\end{table}

This comparison highlights that while PersonaHub excels in generating diverse synthetic personas for broad instruction-following training, \method focuses on preserving the authenticity and depth of established characters for high-fidelity role-playing applications.

\section{Training}
\label{sec:app_training}

\subsection{Training Samples}

We transform conversations from the \method dataset into training samples in the Sharegpt format.
We utilize 90\% of the books in the dataset for training, while the remaining 10\% are set aside to evaluate our models’ ability to generalize to out-of-domain characters and books. 
We construct one training sample for each character in every \method conversation, encompassing both main characters and minor roles.
When training on a character $c$, we designate $c$’s messages as targets for optimization, while using the system prompt and messages from other characters as inputs. Adjacent inputs are concatenated.

Towards general role-playing capabilities across diverse scenarios and applications, we dynamically generate role-playing instructions (system prompts) using varied phrasings, formats, and data types, as shown in Tables ~\ref{tab:prompts_agent} and ~\ref{tab:prompts_agent_2}. 
We consider instructions entirely in natural language, as well as those formatted with special symbols (such as \#\#\#, ===), randomly sampling different formats and various expressions for the same semantics. 
We consider various configurations of the available data. 
Each sample may include (50\%) or exclude (50\%)  the following elements : 
\textit{1)} Profiles of other characters in this conversation; 
\textit{2)} Summaries of the relevant plot;
\textit{3)} Inner thoughts within the messages.

Besides character role-playing, 
we train \method models for environment modeling and next speaker prediction (NSP) for multi-agent simulation. 
For environment modeling, we train LLMs $\agent_e$ to play the environment role $e$ in the same approach, leveraging environment messages in our dataset. 
For NSP, given setting $\mathcal{S}$ and messages $\{m_1,...,m_i\}$, we train LLMs to predict the speaker of $m_{i+1}$ (or ending the conversation).  

Our role-playing dataset comprises approximately 0.1B tokens, as measured using the LLaMA 3.1 tokenizer. 
To maintain general intelligence and instruction-following capabilities, we augment this with an equivalent volume (0.1B tokens) of general-purpose supervised fine-tuning data from Tulu 3~\citep{lambert2024tulu}~\footnote{https://huggingface.co/datasets/allenai/tulu-3-sft-mixture/tree/main/data}. 
This balanced mixture ensures that the model retains broad language understanding while developing specialized role-playing abilities.
If more role-playing data are expected, our data curation pipeline can be easily applied to additional books or other fictional works, thereby acquiring data on a much larger scale.

For more details, please refer to our code.

\subsection{Hyperparameters}

We fine-tune the LLaMA 3.1 models using the following hyperparameters: a learning rate of \(1 \times 10^{-5}\), a sequence length of 16,384, training for 8 epochs, and a global batch size of 48.

\section{Experiment Settings}
\label{sec:exp_setting}

\subsection{Test Set Sampling}

The \method Test set contains 200 samples: 
100 from books used in \method training (in-domain) and 100 from books not used in training (out-of-domain). 
We employ a weighted sampling strategy to prioritize well-established characters with more persona data. The sampling process consists of the following steps:
First, for each book, we assign character weights as the square root of the number of plots in which they are involved.
Second, we calculate the weight of a conversation as the average of its characters’ weights, including both main characters and side roles. Finally, separately for the in-domain and out-of-domain settings, we rank all conversations by weight and perform weighted sampling from the top half of conversations with higher weights.

\subsection{Prompting Strategies for Exisitng RPLA Benchmarks}

For existing RPLA benchmarks, including InCharacter~\citep{wang2024incharacter}, LifeChoice~\citep{xu2024character}, and Cross-MR~\citep{yuan2024evaluating}, we adapt or refine their prompting strategies as follows:

\begin{enumerate}
    \item For InCharacter, we add \dq{\textit{You’re consulting with a personality assessment expert who will ask you some questions. Please provide honest and detailed responses to help with the analysis. Please think carefully and state your reasons when answering the questions.}} after the character profile. 
    This adaptation aims to ensure that RPLAs honestly express their true thoughts. After being trained on authentic character dialogues, the \method models, unlike general LLMs, tend to produce brief, conversational-style answers that may be too short or may decline to answer questions, thus failing to provide sufficient information for  personality assessment.
    \item For LifeChoice and Cross-MR, we reverse the order of their reasoning and answering processes. Specifically, we have them think before providing their choices, thus enabling RPLAs to make well-considered decisions. 
\end{enumerate}

\subsection{Length Correction}
\label{sec:length_correction}

In our evaluation, we use a penalty-based scoring mechanism that counts the flaws in RPLAs' performance. 
However, since longer simulations naturally accumulate more flaws, we need to implement length correction to reduce length bias in LLM judges, following previous work~\citep{li2024judgesurvey}. 

\begin{figure*}[!t]
    \centering
    \includegraphics[width=0.8\textwidth]{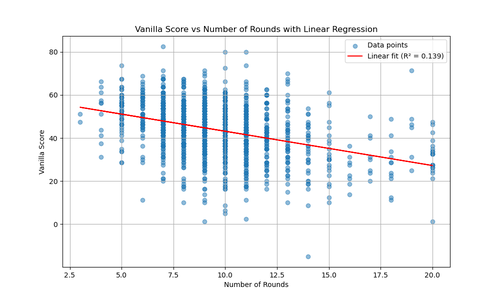}
    \caption{
    Linear regression results showing length bias of penalty-based LLM critics in GCA evaluation.
    }
    \label{fig:length_correction}
\end{figure*}

We analyze the phenomenon of length bias in penalty-based scoring. 
The initial score is defined as \( s = 100-\sum_{f\in\mathcal{F}}v_f \), where \( \mathcal{F} \) represents the set of flaws and \( v_f \) is their severity ranging from 1 to 5. 
Our analysis is conducted on the initial scores from simulations of three models on the \method Test set: \method 70B, LLaMA-3.1 70B, and GPT-4o, with or without retrieval augmentation (three Experience and one Conversation), totaling 1,200 cases. 
We examine the relationship between the number of rounds and the vanilla scores in these 1,200 cases. 
As shown in Figure~\ref{fig:length_correction}, we plot the data points for these cases and perform linear regression to fit these points. 
The fitted linear function is \( \text{score} = -1.5909 \times \text{rounds} + 59.0617 \), which means that for each additional round in the simulation, the score decreases by approximately 1.6 points. 

To mitigate this bias, we implement length correction by compensating  
for the points deducted due to increased rounds. 
Specifically, we compute the length-corrected score as \( s = 100-\sum_{f\in\mathcal{F}}v_f + \lambda|\bar{M}| \), where \( \lambda \) is set to 1.5 based on the analysis above.

\subsection{Human Evaluation}
\label{sec:appendix_human_eval}

The alignment metrics are measured via the frequency of agreement between LLM and human judges when comparing model pairs, excluding cases where judges assign similar scores to both models (\ie, score differences $\leq$ 1 for human judges or $\leq$ 5\% for automatic metrics).

\section{Additional Results}
\label{sec:additional_results}

\paragraph{Evaluation with Different LLMs as Judges}
We conduct additional experiments using alternative judge models for GCA evaluation.
We evaluate 7 representative models as shown in Table~\ref{tab:different_judges},
using three different judge models: GPT-4o, DeepSeek-R1, and DeepSeek-V3. This comparison aims to examine whether our conclusions remain consistent across different LLM judges.

\begin{table}[htbp]
\centering
\renewcommand{\arraystretch}{0.9}
\begin{tabular}{lccc}
\toprule
\textbf{Model} & \textbf{GPT-4o} & \textbf{DeepSeek-R1} & \textbf{DeepSeek-V3} \\
\midrule
GPT-3.5 & 52.8 & 35.9 & 40.5 \\
LLaMA-3.1-8B & 51.8 & 37.2 & 36.8 \\
Abab7-preview & 53.7 & 41.5 & 40.4 \\
\method-8B & 56.1 & 44.5 & 45.9 \\
GPT-4o & \textbf{58.5} & 48.4 & 46.1 \\
Claude-3.5-Sonnet & 56.2 & \textbf{54.8} & 40.7 \\
\method-70B & 57.4 & 50.8 & \textbf{47.7} \\
\bottomrule
\end{tabular}
\caption{Scores (\%) assigned by different LLM judges. Results demonstrate consistent superiority of \method models, while revealing judge-specific preferences.}
\label{tab:different_judges}
\end{table}

The results in Table~\ref{tab:different_judges} reveal several important findings:
\textit{1)} Despite these variations in absolute scores, the relative ranking patterns remain largely consistent, with \method models maintaining competitive or superior performance across all settings.
\textit{2)} Different judges exhibit distinct preferences, confirming potential bias in LLM-based evaluation. While GPT-4o shows preference for GPT models, DeepSeek-R1 and DeepSeek-V3 favor \method-70B and Claude-3.5-Sonnet over GPT-4o.

\paragraph{Win Rates against GPT-4o and GPT-3.5}

To further evaluate existing LLMs in given-circumstance acting, we present their win rates against GPT-3.5~\citep{openai2022chatgpt} and GPT-4o~\citep{OpenAI2023GPT4TR} in Table~\ref{tab:model-comparison-wr}, in addition to the results shown in Table~\ref{tab:model-comparison}.

\begin{table}[htbp]
\centering
\setlength{\tabcolsep}{3pt}
\renewcommand{\arraystretch}{0.9}
{\small
\begin{tabular}{lcc}
\toprule
\multirow{2}{*}{\textbf{Model}} & \multicolumn{2}{c}{\textbf{Win Rate}} \\ \cmidrule(lr){2-3} & \textbf{vs. GPT-3.5} & \textbf{vs. GPT-4o} \\
\midrule
\multicolumn{3}{c}{\textit{Close-source Models}} \\
\midrule
Abab7-preview & 52.17\scriptsize{$\pm4.54$} & 35.33\scriptsize{$\pm2.88$} \\
Doubao-pro & 64.67\scriptsize{$\pm2.32$} & 48.00\scriptsize{$\pm3.97$} \\
Step-1-Flash & 58.75\scriptsize{$\pm6.77$} & 39.75\scriptsize{$\pm1.64$} \\
Step-2 & 65.08\scriptsize{$\pm7.26$} & 46.92\scriptsize{$\pm4.40$} \\
GPT-3.5 & 50.00\scriptsize{$\pm0.00$} & 33.08\scriptsize{$\pm3.64$} \\
GPT-4o & \textbf{66.92\scriptsize{$\pm3.64$}} & \textbf{50.00\scriptsize{$\pm0.00$}} \\
GPT-4o Mini & 59.58\scriptsize{$\pm7.80$} & 42.92\scriptsize{$\pm0.63$} \\
Gemini Pro & 66.17\scriptsize{$\pm3.33$} & 49.75\scriptsize{$\pm0.25$} \\
Claude-3-Haiku & 51.42\scriptsize{$\pm4.40$} & 33.25\scriptsize{$\pm3.03$} \\
Claude-3.5-Sonnet & 58.17\scriptsize{$\pm8.00$} & 41.42\scriptsize{$\pm1.77$} \\
\midrule
\multicolumn{3}{c}{\textit{Open-source Models}} \\
\midrule
Mistral-7B & 41.50\scriptsize{$\pm7.28$} & 26.67\scriptsize{$\pm1.81$} \\
Qwen-2-7B & 24.83\scriptsize{$\pm6.64$} & 13.58\scriptsize{$\pm3.00$} \\
LLaMA-3.1-8B & 45.33\scriptsize{$\pm6.57$} & 32.75\scriptsize{$\pm5.86$} \\
CoSER-8B & 58.17\scriptsize{$\pm5.36$} & 41.67\scriptsize{$\pm3.75$} \\
Vicuna-13B-1.5 & 30.50\scriptsize{$\pm10.44$} & 18.67\scriptsize{$\pm2.18$} \\
Mixtral-8x7B & 32.50\scriptsize{$\pm1.50$} & 19.83\scriptsize{$\pm1.42$} \\
Qwen-2-72B & 59.58\scriptsize{$\pm5.11$} & 41.92\scriptsize{$\pm1.66$} \\
LLaMA-3.1-70B & 53.83\scriptsize{$\pm2.57$} & 37.92\scriptsize{$\pm3.26$} \\
Higgs-Llama-3-70B & 51.83\scriptsize{$\pm4.89$} & 34.92\scriptsize{$\pm4.84$} \\
CoSER-70B & \underline{64.33\scriptsize{$\pm3.74$}} & \underline{49.42\scriptsize{$\pm4.11$}} \\
DeepSeek-V3 & 59.50\scriptsize{$\pm2.82$} & 39.50\scriptsize{$\pm2.05$} \\
\bottomrule
\end{tabular}}
\caption{Win rates (\%) of various LLMs on given-circumstance acting using \method Test, compared against GPT-3.5 and GPT-4. \textbf{Bold} and \underline{underlined} numbers indicate the best results among all and open-source models, respectively.}
\label{tab:model-comparison-wr}
\end{table}

\paragraph{Generalization of \method Models to New Characters}

Table~\ref{tab:exp_idood} separately presents the performance of LLMs on test splits from books included in and excluded from \method training. 
The results demonstrate consistent trends across both splits, confirming that \method models maintain strong performance even on out-of-domain characters.

\begin{table*}[htbp]
\centering
\setlength{\tabcolsep}{2pt}
\renewcommand{\arraystretch}{0.9}
\resizebox{\linewidth}{!}{\small
\begin{tabular}{lcccccccccc}
\toprule
\multirow{3}{*}{\textbf{Model}} & \multicolumn{5}{c}{\textbf{In-Domain}} & \multicolumn{5}{c}{\textbf{Out-of-Domain}} \\
\cmidrule(lr){2-6}
\cmidrule(lr){7-11}
\multirow{2}{*}{} & \multirow{2}{*}{\makecell{\scriptsize \textbf{Storyline}\\ \scriptsize  \textbf{Consistency}}} & \multirow{2}{*}{\makecell{\scriptsize \textbf{Anthro-}\\ \scriptsize  \textbf{pomorphism}}} & \multirow{2}{*}{\makecell{\scriptsize \textbf{Character}\\ \scriptsize  \textbf{Fidelity}}} & \multirow{2}{*}{\makecell{\scriptsize \textbf{Storyline}\\ \scriptsize  \textbf{Quality}}} & \multirow{2}{*}{\makecell{\scriptsize \textbf{Average}\\ \scriptsize  \textbf{Score}}} & \multirow{2}{*}{\makecell{\scriptsize \textbf{Storyline}\\ \scriptsize  \textbf{Consistency}}} & \multirow{2}{*}{\makecell{\scriptsize \textbf{Anthro-}\\ \scriptsize  \textbf{pomorphism}}} & \multirow{2}{*}{\makecell{\scriptsize \textbf{Character}\\ \scriptsize  \textbf{Fidelity}}} & \multirow{2}{*}{\makecell{\scriptsize \textbf{Storyline}\\ \scriptsize  \textbf{Quality}}} & \multirow{2}{*}{\makecell{\scriptsize \textbf{Average}\\ \scriptsize  \textbf{Score}}} \\
 &  &  &  &  &  &  &  &  &  & \\
\midrule
\multicolumn{11}{c}{\textit{Close-source Models}} \\
\midrule
Abab7-preview & 55.31\tiny{$\pm1.4$} & 42.29\tiny{$\pm1.3$} & 42.94\tiny{$\pm3.2$} & 74.13\tiny{$\pm1.9$} & 53.67\tiny{$\pm1.0$} & 58.30\tiny{$\pm1.8$} & 46.17\tiny{$\pm2.6$} & 44.72\tiny{$\pm2.3$} & 75.54\tiny{$\pm1.5$} & 56.18\tiny{$\pm0.1$}\\
Doubao-pro & 60.37\tiny{$\pm1.8$} & 49.06\tiny{$\pm0.4$} & 45.76\tiny{$\pm3.4$} & 77.87\tiny{$\pm1.2$} & 58.26\tiny{$\pm0.9$} & 61.53\tiny{$\pm1.1$} & 50.38\tiny{$\pm0.3$} & 48.28\tiny{$\pm1.7$} & 80.69\tiny{$\pm0.4$} & 60.22\tiny{$\pm0.4$}\\
Step-1-Flash & 57.10\tiny{$\pm0.3$} & 48.31\tiny{$\pm1.0$} & 41.84\tiny{$\pm1.9$} & 76.18\tiny{$\pm1.5$} & 55.86\tiny{$\pm1.0$} & 58.39\tiny{$\pm1.2$} & 47.94\tiny{$\pm0.5$} & 47.12\tiny{$\pm1.0$} & 75.67\tiny{$\pm0.5$} & 57.28\tiny{$\pm0.2$}\\
Step-2 & \textbf{60.55\tiny{$\pm0.7$}} & 48.82\tiny{$\pm2.2$} & \textbf{47.90\tiny{$\pm1.1$}} & 77.33\tiny{$\pm1.0$} & \textbf{58.65\tiny{$\pm1.0$}} & \textbf{62.30\tiny{$\pm1.5$}} & 49.30\tiny{$\pm1.2$} & \textbf{46.77\tiny{$\pm1.7$}} & 78.59\tiny{$\pm1.2$} & \textbf{59.24\tiny{$\pm0.6$}}\\
GPT-3.5 & 55.58\tiny{$\pm0.9$} & 42.18\tiny{$\pm5.2$} & 40.37\tiny{$\pm3.2$} & 72.90\tiny{$\pm0.1$} & 52.76\tiny{$\pm2.0$} & 59.69\tiny{$\pm3.0$} & 44.25\tiny{$\pm2.1$} & 44.60\tiny{$\pm1.4$} & 74.66\tiny{$\pm2.3$} & 55.80\tiny{$\pm1.6$}\\
GPT-4o & 59.88\tiny{$\pm1.4$} & 48.11\tiny{$\pm1.0$} & 47.10\tiny{$\pm0.2$} & \textbf{79.06\tiny{$\pm1.5$}} & 58.54\tiny{$\pm0.7$} & 62.29\tiny{$\pm1.5$} & 49.48\tiny{$\pm1.2$} & 49.90\tiny{$\pm0.4$} & \textbf{80.18\tiny{$\pm0.9$}} & 60.46\tiny{$\pm0.1$}\\
GPT-4o Mini & 59.15\tiny{$\pm1.3$} & 46.94\tiny{$\pm1.5$} & 43.99\tiny{$\pm2.5$} & 77.33\tiny{$\pm0.6$} & 56.85\tiny{$\pm0.1$} & 61.02\tiny{$\pm1.6$} & 49.48\tiny{$\pm3.1$} & 45.77\tiny{$\pm1.1$} & 79.77\tiny{$\pm0.5$} & 59.01\tiny{$\pm1.4$}\\
Gemini Pro & 57.72\tiny{$\pm0.4$} & 50.94\tiny{$\pm1.8$} & 46.23\tiny{$\pm1.0$} & 76.22\tiny{$\pm1.6$} & 57.78\tiny{$\pm0.9$} & 60.50\tiny{$\pm1.9$} & 53.88\tiny{$\pm1.1$} & 49.43\tiny{$\pm0.3$} & 78.97\tiny{$\pm1.3$} & 60.69\tiny{$\pm0.8$}\\
Claude-3-Haiku & 57.61\tiny{$\pm0.5$} & 44.97\tiny{$\pm2.2$} & 40.61\tiny{$\pm1.4$} & 73.52\tiny{$\pm1.2$} & 54.18\tiny{$\pm0.6$} & 58.74\tiny{$\pm1.1$} & 44.36\tiny{$\pm1.9$} & 43.14\tiny{$\pm0.8$} & 74.76\tiny{$\pm1.7$} & 55.25\tiny{$\pm1.2$}\\
Claude-3.5-Sonnet & 56.44\tiny{$\pm1.5$} & 47.24\tiny{$\pm1.4$} & 44.89\tiny{$\pm0.6$} & 76.39\tiny{$\pm1.5$} & 56.24\tiny{$\pm1.0$} & 58.46\tiny{$\pm1.1$} & 49.75\tiny{$\pm3.3$} & 46.49\tiny{$\pm3.0$} & 78.06\tiny{$\pm1.5$} & 58.19\tiny{$\pm0.9$}\\
\midrule
\multicolumn{11}{c}{\textit{Open-source Models}} \\
\midrule
Mistral-7B & 60.29\tiny{$\pm1.9$} & 38.98\tiny{$\pm2.0$} & 42.93\tiny{$\pm0.5$} & 62.20\tiny{$\pm3.1$} & 51.10\tiny{$\pm1.8$} & 59.51\tiny{$\pm2.4$} & 41.02\tiny{$\pm1.3$} & 46.57\tiny{$\pm2.8$} & 61.65\tiny{$\pm1.3$} & 52.19\tiny{$\pm0.7$}\\
Qwen-2-7B & 50.77\tiny{$\pm1.3$} & 34.17\tiny{$\pm1.1$} & 29.92\tiny{$\pm2.9$} & 62.58\tiny{$\pm0.7$} & 44.36\tiny{$\pm0.7$} & 53.14\tiny{$\pm1.5$} & 36.79\tiny{$\pm1.0$} & 33.09\tiny{$\pm3.3$} & 63.79\tiny{$\pm1.4$} & 46.70\tiny{$\pm1.3$}\\
LLaMA-3.1-8B & 53.00\tiny{$\pm1.2$} & 43.64\tiny{$\pm0.8$} & 39.05\tiny{$\pm1.5$} & 71.34\tiny{$\pm1.3$} & 51.76\tiny{$\pm0.6$} & 55.20\tiny{$\pm2.1$} & 47.08\tiny{$\pm3.2$} & 41.38\tiny{$\pm1.0$} & 73.23\tiny{$\pm2.5$} & 54.22\tiny{$\pm1.8$}\\
CoSER-8B & 58.56\tiny{$\pm3.5$} & 46.78\tiny{$\pm1.1$} & 45.78\tiny{$\pm3.1$} & 73.38\tiny{$\pm1.8$} & 56.12\tiny{$\pm0.5$} & 58.66\tiny{$\pm1.6$} & 47.69\tiny{$\pm0.8$} & 48.03\tiny{$\pm1.3$} & 72.71\tiny{$\pm1.2$} & 56.77\tiny{$\pm0.6$}\\
Vicuna-13B-1.5 & 51.84\tiny{$\pm1.2$} & 38.88\tiny{$\pm0.4$} & 36.39\tiny{$\pm0.5$} & 58.15\tiny{$\pm0.7$} & 46.31\tiny{$\pm0.4$} & 53.66\tiny{$\pm2.1$} & 39.35\tiny{$\pm2.7$} & 39.69\tiny{$\pm1.6$} & 62.71\tiny{$\pm2.5$} & 48.85\tiny{$\pm2.1$}\\
Mixtral-8x7B & 51.18\tiny{$\pm3.1$} & 38.76\tiny{$\pm1.8$} & 34.37\tiny{$\pm2.8$} & 66.44\tiny{$\pm0.1$} & 47.69\tiny{$\pm1.7$} & 51.32\tiny{$\pm0.4$} & 38.13\tiny{$\pm0.5$} & 39.48\tiny{$\pm2.6$} & 68.93\tiny{$\pm1.7$} & 49.47\tiny{$\pm1.1$}\\
Qwen-2-72B & 56.34\tiny{$\pm2.4$} & 46.19\tiny{$\pm0.4$} & 45.59\tiny{$\pm2.2$} & 75.68\tiny{$\pm0.3$} & 55.95\tiny{$\pm1.0$} & 59.15\tiny{$\pm1.1$} & 48.37\tiny{$\pm1.3$} & 47.65\tiny{$\pm1.8$} & 77.52\tiny{$\pm0.4$} & 58.17\tiny{$\pm1.0$}\\
LLaMA-3.1-70B & 55.44\tiny{$\pm2.7$} & 44.19\tiny{$\pm1.9$} & 42.67\tiny{$\pm1.5$} & 73.90\tiny{$\pm1.8$} & 54.05\tiny{$\pm0.8$} & 59.48\tiny{$\pm1.4$} & 47.72\tiny{$\pm2.3$} & 44.78\tiny{$\pm0.9$} & 75.78\tiny{$\pm0.9$} & 56.94\tiny{$\pm0.7$}\\
Higgs-Llama-3-70B & 55.85\tiny{$\pm2.7$} & 41.18\tiny{$\pm3.3$} & 39.79\tiny{$\pm2.3$} & 73.77\tiny{$\pm0.6$} & 52.65\tiny{$\pm2.0$} & 58.35\tiny{$\pm0.5$} & 46.45\tiny{$\pm1.1$} & 45.04\tiny{$\pm1.0$} & 77.48\tiny{$\pm0.6$} & 56.83\tiny{$\pm0.6$}\\
CoSER-70B & \underline{57.77\tiny{$\pm1.3$}} & \underline{\textbf{51.60\tiny{$\pm1.3$}}} & \underline{45.82\tiny{$\pm0.9$}} & 74.27\tiny{$\pm1.1$} & \underline{57.37\tiny{$\pm0.7$}} & \underline{59.56\tiny{$\pm2.1$}} & \underline{\textbf{55.06\tiny{$\pm1.1$}}} & \underline{51.67\tiny{$\pm2.5$}} & 76.71\tiny{$\pm1.6$} & \underline{60.75\tiny{$\pm1.1$}}\\
DeepSeek-V3 & 55.36\tiny{$\pm0.5$} & 47.55\tiny{$\pm1.3$} & 43.10\tiny{$\pm0.3$} & \underline{74.91\tiny{$\pm2.0$}} & 55.23\tiny{$\pm0.6$} & 57.45\tiny{$\pm2.0$} & 48.19\tiny{$\pm0.9$} & 44.93\tiny{$\pm0.4$} & \underline{78.41\tiny{$\pm1.1$}} & 57.24\tiny{$\pm0.5$}\\
\bottomrule
\end{tabular}}
\caption{LLM performance (\%) on given-circumstance acting using \method Test, separated into the in-domain and out-of-domain splits for \method training.}
\label{tab:exp_idood}
\end{table*}

\paragraph{Conversation Continuation}

Table ~\ref{tab:model-comparison-cf3} %
shows experiment results when multi-agent systems continue conversations from the first $k=3$ original messages.  

Tables ~\ref{tab:model-comparison-cf3} presents detailed evaluation results where our multi-agent simulations start from the first $k=3$ original messages.

\begin{table*}[htbp]
\centering
\setlength{\tabcolsep}{3pt}
\renewcommand{\arraystretch}{0.9}
\resizebox{0.9\linewidth}{!}{
\begin{tabular}{lccccccc}
\toprule
\multirow{3}{*}{\textbf{Model}} & \multicolumn{5}{c}{\textbf{Based on LLM Judges}} & \multicolumn{2}{c}{\textbf{Based on N-gram}} \\ \cmidrule(lr){2-6} \cmidrule(lr){7-8} & \multirow{2}{*}{\makecell{\textbf{Storyline}\\\textbf{Consistency}}} & \multirow{2}{*}{\makecell{\textbf{Anthro-}\\\textbf{pomorphism}}} & \multirow{2}{*}{\makecell{\textbf{Character}\\\textbf{Fidelity}}} & \multirow{2}{*}{\makecell{\textbf{Storyline}\\\textbf{Quality}}} & \multirow{2}{*}{\makecell{\textbf{Average}\\\textbf{Score}}} & \multirow{2}{*}{\makecell{\textbf{BLEU}}} & \multirow{2}{*}{\makecell{\textbf{ROUGE-L}}} \\
 &  &  &  &  &  &  &  \\
\midrule
\multicolumn{8}{c}{\textit{Close-source Models}} \\
\midrule
Abab7-preview & 65.25 & 55.26 & 55.95 & 79.68 & 64.03 & 10.53 & 15.99 \\
Doubao-pro & \textbf{68.31} & \textbf{59.51} & 59.23 & 80.15 & 66.80 & 11.83 & 17.13 \\
Step-1-Flash & 64.32 & 53.35 & 54.38 & 79.03 & 62.77 & 11.94 & 17.06 \\
Step-2 & 66.61 & 55.81 & 59.59 & 80.56 & 65.64 & 11.72 & 17.27 \\
GPT-3.5 & 65.72 & 54.34 & 56.48 & 77.67 & 63.55 & 10.80 & 16.39 \\
GPT-4o & 67.48 & 58.51 & 62.30 & \textbf{82.23} & \textbf{67.63} & 12.32 & 16.98 \\
GPT-4o Mini & 66.74 & 55.55 & 56.80 & 80.01 & 64.77 & 9.50 & 15.54 \\
Gemini Pro & 65.47 & 59.43 & \textbf{62.42} & 78.78 & 66.53 & 10.84 & 16.27 \\
Claude-3-Haiku & 64.51 & 54.01 & 57.13 & 77.26 & 63.23 & 10.11 & 16.18 \\
Claude-3.5-Sonnet & 64.54 & 54.57 & 58.76 & 79.89 & 64.44 & 8.64 & 14.94 \\
\midrule
\multicolumn{8}{c}{\textit{Open-source Models}} \\
\midrule
Mistral-7B & \underline{67.50} & 50.39 & 59.90 & 68.67 & 61.62 & 7.41 & 14.33 \\
Qwen-2-7B & 59.81 & 49.87 & 49.04 & 69.16 & 56.97 & 10.59 & 16.22 \\
LLaMA-3.1-8B & 60.90 & 51.36 & 50.37 & 74.89 & 59.38 & 7.86 & 13.82 \\
CoSER-8B & 67.22 & 58.19 & 58.80 & 76.44 & 65.16 & 13.17 & \underline{\textbf{18.42}} \\
Vicuna-13B-1.5 & 60.92 & 46.68 & 50.51 & 67.80 & 56.48 & 5.11 & 9.71 \\
Mixtral-8x7B & 64.66 & 51.21 & 54.08 & 74.01 & 60.99 & 11.21 & 16.97 \\
Qwen-2-72B & 67.27 & 55.87 & 59.84 & \underline{80.07} & \underline{65.76} & 11.92 & 16.96 \\
LLaMA-3.1-70B & 64.08 & 54.80 & 54.18 & 78.31 & 62.84 & 8.74 & 14.74 \\
Higgs-Llama-3-70B & 65.09 & 54.80 & 58.20 & 79.36 & 64.36 & 10.86 & 16.40 \\
CoSER-70B & 65.99 & \underline{59.24} & \underline{59.97} & 76.74 & 65.48 & \underline{\textbf{13.46}} & 18.18 \\
DeepSeek-V3 & 62.95 & 56.95 & 58.25 & 79.80 & 64.49 & 9.25 & 15.01 \\
\bottomrule
\end{tabular}}
\caption{Performance (\%) of various LLMs on \method Test  in conversation continuation setting ($k=3$), where RPLAs start from the first three messages in the authentic conversations.}
\label{tab:model-comparison-cf3}
\end{table*}

\paragraph{\method Dataset for Retrieval Augmentation}

We validate the effectiveness of \method’s comprehensive data types for retrieval augmentation on the \method Test set. 
We explore three retrieval sources related to specific characters: dialogues from conversations, experiences from plots, and raw text from plots. 
We compare several combinations of these sources, including:
\textit{1)} None (Base)
\textit{2)} Raw text of one plot (Raw Text)
\textit{3)} One conversation (Conv.)
\textit{4)} Character experiences from three plots (Expr.3)
\textit{5)} Expr.3 combined with Conv.
\textit{6)} Expr.10 combined with Conv.
The complete results are presented in Table~\ref{tab:rag}.

\paragraph{Ablation Studies}

We examine the effectiveness of inner thoughts in both training and evaluation. 
The complete results are demonstrated in Table~\ref{tab:model-comparison-wo-cot-full}.

\begin{table*}[htbp]
\centering
\setlength{\tabcolsep}{3pt}
\renewcommand{\arraystretch}{0.9}
\resizebox{0.9\textwidth}{!}{
\begin{tabular}{lccccccc}
\toprule
\multirow{3}{*}{\textbf{Model}} & \multicolumn{5}{c}{\textbf{Based on LLM Judges}} & \multicolumn{2}{c}{\textbf{Based on N-gram}} \\ \cmidrule(lr){2-6} \cmidrule(lr){7-8} & \multirow{2}{*}{\makecell{\textbf{Storyline}\\\textbf{Consistency}}} & \multirow{2}{*}{\makecell{\textbf{Anthro-}\\\textbf{pomorphism}}} & \multirow{2}{*}{\makecell{\textbf{Character}\\\textbf{Fidelity}}} & \multirow{2}{*}{\makecell{\textbf{Storyline}\\\textbf{Quality}}} & \multirow{2}{*}{\makecell{\textbf{Average}\\\textbf{Score}}} & \multirow{2}{*}{\makecell{\textbf{BLEU}}} & \multirow{2}{*}{\makecell{\textbf{ROUGE-L}}} \\
 &  &  &  &  &  &  &  \\
\midrule
\multicolumn{8}{c}{\textit{Test w/ I.T. }} \\
\midrule
GPT-4o & 61.59 & 48.93 & 48.95 & 80.33 & 59.95 & 5.90 & 12.11 \\
Qwen-2-72B & 57.75 & 47.28 & 46.62 & 76.60 & 57.06 & 5.38 & 11.85 \\
LLaMA-3.1-70B & 57.46 & 45.95 & 43.72 & 74.84 & 55.49 & 4.82 & 10.98 \\
CoSER-70B & 58.66 & 53.33 & 48.75 & 75.49 & 59.06 & 10.10 & 14.78 \\
 \hfill\hfill\hfill {\textit{trained w/o} \textsc{I.T.}} & 55.67 & 49.93 & 45.67 & 72.89 & 56.04 & 9.18 & 14.18 \\
LLaMA-3.1-8B & 54.10 & 45.36 & 40.22 & 72.29 & 52.99 & 4.59 & 10.18 \\
CoSER-8B & 58.61 & 47.23 & 46.90 & 73.04 & 56.45 & 9.40 & 14.21 \\
 \hfill\hfill\hfill {\textit{trained w/o} \textsc{I.T.}} & 54.91 & 44.89 & 44.10 & 73.09 & 54.25 & 9.65 & 14.27 \\
\midrule
\multicolumn{8}{c}{\textit{Test w/o I.T. }} \\
\midrule
GPT-4o & 59.51 & 45.86 & 45.13 & 77.06 & 56.89 & 5.54 & 11.44 \\
Qwen-2-72B & 55.39 & 41.00 & 37.70 & 73.70 & 51.95 & 5.22 & 11.32 \\
LLaMA-3.1-70B & 56.01 & 43.17 & 41.23 & 72.08 & 53.12 & 4.15 & 10.37 \\
CoSER-70B & 59.23 & 48.76 & 47.06 & 74.23 & 57.32 & 10.02 & 14.72 \\
 \hfill\hfill\hfill {\textit{trained w/o} \textsc{I.T.}} & 55.52 & 46.22 & 45.34 & 74.31 & 55.34 & 9.66 & 14.62 \\
LLaMA-3.1-8B & 54.27 & 45.54 & 37.86 & 70.23 & 51.97 & 4.37 & 10.27 \\
CoSER-8B & 56.78 & 45.64 & 45.00 & 71.16 & 54.65 & 9.38 & 14.44 \\
 \hfill\hfill\hfill {\textit{trained w/o} \textsc{I.T.}} & 54.98 & 46.22 & 45.81 & 70.52 & 54.38 & 9.16 & 13.97 \\
\bottomrule
\end{tabular}}
\caption{Comprehensive results of LLM performance (\%) on \method Test, with and without inner thoughts (I.T) during inference and \method model training.}
\label{tab:model-comparison-wo-cot-full}
\end{table*}

\begin{table*}[htbp]
\centering
\setlength{\tabcolsep}{3pt}
\renewcommand{\arraystretch}{0.9}
\resizebox{0.9\textwidth}{!}{
\begin{tabular}{lccccccc}
\toprule
\multirow{3}{*}{\textbf{Augmented By}} & \multicolumn{5}{c|}{\textbf{Based on LLM Judges}} & \multicolumn{2}{c}{\textbf{Based on N-gram}} \\ \cmidrule(lr){2-6} \cmidrule(lr){7-8} & \multirow{2}{*}{\makecell{\textbf{Storyline}\\\textbf{Consistency}}} & \multirow{2}{*}{\makecell{\textbf{Anthro-}\\\textbf{pomorphism}}} & \multirow{2}{*}{\makecell{\textbf{Character}\\\textbf{Fidelity}}} & \multirow{2}{*}{\makecell{\textbf{Storyline}\\\textbf{Quality}}} & \multirow{2}{*}{\makecell{\textbf{Average}\\\textbf{Score}}} & \multirow{2}{*}{\makecell{\textbf{BLEU}}} & \multirow{2}{*}{\makecell{\textbf{ROUGE-L}}} \\
 &  &  &  &  &  &  &  \\
\midrule
\multicolumn{8}{c}{\textbf{GPT-4o}} \\
\midrule
Base & \underline{61.59} & 48.93 & 48.95 & \textbf{80.33} & \underline{59.95} & 5.90 & 12.11 \\
Raw Text & 59.60 & 48.23 & 48.83 & 80.03 & 59.17 & 5.83 & 12.08 \\
Conv. & 60.03 & \textbf{50.99} & 47.11 & 78.03 & 59.04 & \textbf{7.20} & \textbf{12.77} \\
Expr.3 & \textbf{61.98} & 48.94 & 48.56 & 79.93 & 59.85 & 5.81 & 12.13 \\
Expr.3 + Conv. & 59.81 & \underline{50.11} & \textbf{51.12} & \underline{80.29} & \textbf{60.33} & \underline{7.05} & \underline{12.67} \\
Expr.10 + Conv. & 60.99 & 48.70 & \underline{49.89} & 79.22 & 59.70 & 6.95 & 12.42 \\
\midrule
\multicolumn{8}{c}{\textbf{LLaMA-3.1-8B}} \\
\midrule
Base & 54.10 & 45.36 & 40.22 & 72.29 & 52.99 & 4.59 & 10.18 \\
Raw Text & 55.41 & 45.37 & 41.66 & 74.31 & 54.19 & 4.73 & 10.51 \\
Conv. & 56.54 & 45.34 & 40.37 & 71.74 & 53.50 & 5.19 & 10.67 \\
Expr.3 & \underline{56.78} & 46.70 & 41.18 & 72.38 & 54.26 & 4.67 & 10.30 \\
Expr.3 + Conv. & \textbf{57.36} & \textbf{47.92} & \textbf{42.60} & \underline{74.60} & \textbf{55.62} & \underline{5.21} & \underline{10.81} \\
Expr.10 + Conv. & 56.41 & \underline{47.18} & \underline{42.09} & \textbf{74.74} & \underline{55.10} & \textbf{5.64} & \textbf{11.11} \\
\midrule
\multicolumn{8}{c}{\textbf{LLaMA-3.1-70B}} \\
\midrule
Base & 57.46 & \underline{45.95} & 43.72 & 74.84 & 55.49 & 4.82 & 10.98 \\
Raw Text & \underline{57.52} & \textbf{47.46} & 43.75 & 74.06 & \textbf{55.70} & 4.90 & 10.97 \\
Conv. & 54.93 & 44.93 & \textbf{46.21} & \textbf{76.53} & \underline{55.65} & 5.45 & 11.13 \\
Expr.3 & 56.59 & 45.03 & 42.59 & 74.70 & 54.73 & 4.78 & 10.82 \\
Expr.3 + Conv. & 57.36 & 44.89 & 43.49 & \underline{74.86} & 55.15 & \textbf{5.58} & \underline{11.14} \\
Expr.10 + Conv. & \textbf{58.05} & 45.81 & \underline{44.47} & 74.01 & 55.59 & \underline{5.53} & \textbf{11.33} \\
\midrule
\multicolumn{8}{c}{\textbf{CoSER-8B}} \\
\midrule
Base & 58.61 & 47.23 & 46.90 & 73.04 & 56.45 & 9.40 & 14.21 \\
Raw Text & 57.64 & 45.25 & 44.27 & 72.04 & 54.80 & 9.34 & 14.41 \\
Conv. & 57.99 & 46.66 & 47.32 & 73.99 & 56.49 & \underline{13.64} & \underline{18.15} \\
Expr.3 & \underline{59.84} & \underline{47.41} & 48.16 & 74.51 & \underline{57.48} & 9.37 & 14.34 \\
Expr.3 + Conv. & \textbf{59.94} & 45.25 & \underline{48.52} & \underline{74.96} & 57.17 & \textbf{13.90} & \textbf{18.57} \\
Expr.10 + Conv. & 58.34 & \textbf{48.14} & \textbf{48.72} & \textbf{75.28} & \textbf{57.62} & 13.49 & 17.85 \\
\midrule
\multicolumn{8}{c}{\textbf{CoSER-70B}} \\
\midrule
Base & 58.66 & 53.33 & 48.75 & 75.49 & 59.06 & 10.10 & 14.78 \\
Raw Text & 60.64 & 52.85 & 47.97 & 74.02 & 58.87 & 10.54 & 15.41 \\
Conv. & \textbf{64.59} & \textbf{53.79} & \textbf{54.86} & 77.28 & \textbf{62.63} & \textbf{17.22} & \textbf{21.17} \\
Expr.3 & 58.67 & 52.69 & 50.66 & 74.67 & 59.17 & 10.00 & 14.82 \\
Expr.3 + Conv. & \underline{61.58} & \underline{53.78} & \underline{52.00} & \underline{77.47} & \underline{61.21} & \underline{15.98} & \underline{19.95} \\
Expr.10 + Conv. & 61.53 & 52.58 & 50.80 & \textbf{78.07} & 60.75 & 15.80 & 19.90 \\
\midrule
\multicolumn{8}{c}{\textbf{Qwen-2-72B}} \\
\midrule
Base & 57.75 & 47.28 & \textbf{46.62} & 76.60 & 57.06 & 5.38 & 11.85 \\
Raw Text & 58.89 & \underline{47.31} & 45.28 & 76.78 & 57.06 & 5.07 & 11.57 \\
Conv. & 59.11 & \textbf{47.82} & 44.19 & \underline{77.54} & 57.16 & 5.77 & 12.05 \\
Expr.3 & 59.17 & 47.21 & 44.85 & 77.47 & 57.18 & 5.45 & 11.94 \\
Expr.3 + Conv. & \underline{59.91} & 46.89 & 46.06 & \textbf{79.31} & \textbf{58.04} & \textbf{6.17} & \textbf{12.25} \\
Expr.10 + Conv. & \textbf{60.14} & 46.84 & \underline{46.27} & 76.82 & \underline{57.52} & \underline{5.85} & \underline{12.19} \\
\bottomrule
\end{tabular}}
\caption{
Comprehensive results of LLM performance (\%) on \method Test with retrieval augmentation from various character data. Expr. and Conv. denote experiences and conversations.  
\textbf{Bold} and \underline{underlined} numbers denote the best and second-best results among different retrieval settings, respectively. 
}
\label{tab:rag}
\end{table*}

\paragraph{Comparing GCA Simulation with AI-driven Storytelling Methods}

To provide a more comprehensive evaluation against prior work in digital actors and AI-driven storytelling, we compare our GCA simulation approach with HollmWood~\citep{chen-etal-2024-hollmwood}, a representative method for AI-generated storytelling. 
We conduct this comparison using 30 samples from \method Test, employing GPT-4o as both the actor and writer in both frameworks to ensure fair comparison.
The generated stories and character interactions are evaluated using GCA evaluation.
Results demonstrate that GCA simulation achieves an average score of 59.2\%, significantly outperforming HollmWood's 50.2\%.
These findings highlight that our given-circumstance acting approach produces more authentic and human-like character interactions compared to previous AI storytelling methods.
The superior performance of GCA simulation can be attributed to its focus on character-driven dialogue generation within well-defined circumstances, rather than plot-driven narrative construction, leading to more faithful character portrayals and realistic conversational dynamics.

\section{Authentic v.s. LLM-Generated Role-playing Data}

To demonstrate the superiority of authentic role-playing data over synthesized alternatives, we conduct two complementary experiments comparing authentic conversations from \method dataset with LLM-generated dialogues.

\paragraph{GCA Evaluation of Dialogue Quality}
We apply GCA evaluation to score authentic (groundtruth) and LLM-generated dialogues on 100 samples from \method Test without providing groundtruth to LLM judges as reference. 
We use DeepSeek-R1 as the judge model due to its superior alignment with human evaluators (as shown in \S\ref{sec:human_eval}).
GPT-4o is used to generate synthetic dialogues via GCA simulation.
Results show that authentic dialogues from \method achieve a score of 85.1\%, significantly outperforming GPT-4o generated dialogues at 76.2\%.

\paragraph{Training Effectiveness Comparison}
We further examine the training effectiveness of authentic versus synthetic dialogues. 
We fine-tune LLaMA-3.1-8B using authentic or LLM-generated dialogues for 200 samples from \method Test for 4 epochs, using the same settings as \method-8B. 
We evaluate the two trained models via GCA evaluation on 100 new samples. 
The model trained on authentic dialogues from \method achieves 55.3\% score, while the model trained on GPT-4o generated dialogues achieves 54.7\%.

These results demonstrate that authentic dialogues consistently outperform LLM-generated alternatives in both intrinsic quality and training effectiveness.
These findings highlight the importance of authentic, high-quality role-playing data for developing effective RPLAs.

\section{Examples and Case Study}

We present several examples of our extracted conversations, as well as corresponding  simulations in this given circumstance by LLMs.

Tables~\ref{tab:case_sansa} to~\ref{tab:case_sansa_2} illustrate a classic  conversation extracted from \textit{A Storm of Swords (A Song of Ice and Fire, \#3)} and the corresponding simulation by \method 70B. In \method 70B’s simulation, when confronted with \textit{Sansa Stark}, \textit{Lysa Arryn} utters her iconic line, \textit{“Grown enough to be wed, wed enough to be bedded”}, reflecting her personality and worldview. 
This indicates that \method models excellently recall and apply character-related knowledge from their pretrained data.

Tables \ref{tab:case_cerci} to \ref{tab:case_cerci4} present another scene from \textit{A Dance with Dragons (A Song of Ice and Fire, \#5)}, specifically the \textit{walk of atonement}, in which \textit{Cersei Lannister} is forced to walk naked through the streets, facing both physical and mental humiliation while striving to preserve her dignity. 
We present the original dialogue alongside simulations by \method 70B, GPT-4o, and Claude-3.5-Sonnet. 
Notably, \method 70B faithfully captures the suppressed anger of \textit{Cersei Lannister} as depicted in the original conversation, whereas the other models, including GPT-4o and Claude-3.5-Sonnet, resort to a stereotypical portrayal of her arrogance and pride.

Table~\ref{tab:critique_example} presents an output example of LLM critics, demonstrating the identified performance flaws in GCA simulation. 

\begin{table*}[h]
\centering
\resizebox{\linewidth}{!}{
\small
\begin{tabular}{p{1in}|p{5.4in}}
\toprule
\multicolumn{2}{c}{\textbf{Simulation Case (from \textit{A Storm of Swords (A Song of Ice and Fire, \#3))}}} \\
\midrule

\textbf{Plot Summary} & Sansa is confronted by her jealous aunt Lysa, who accuses her of kissing Petyr Baelish. Lysa threatens to push Sansa through the Moon Door, but Petyr intervenes. He calms Lysa down, only to push her through the Moon Door himself, killing her.
    \\ \midrule

{\textbf{Scenario}} &
    In the High Hall of the Eyrie, Lysa Arryn confronts Sansa about kissing Petyr Baelish, becoming increasingly unstable and dangerous ...
    \\ \midrule

{\textbf{Characters}} &
    {\large{\textbf{\textit{{Sansa Stark}}}}}

    \textbf{\textit{{Motivation}}}: I'm terrified. Aunt Lysa has gone mad. I need to find a way out of this situation before she hurts me.

    \textbf{\textit{{Profile}}}: Sansa Stark, a young noblewoman from the North, finds herself at the center of political intrigue and personal tragedy in "A Storm of Swords." Born to the noble House Stark, Sansa begins the story as a ward of the crown in King's Landing, her initial romantic notions of court life long since shattered. Forced into an unwanted marriage with Tyrion Lannister, Sansa endures the ceremony and subsequent feast with a mix of reluctance, sadness, and carefully hidden disgust ...

    \quad 
    
    {\large{\textbf{\textit{{Lysa Arryn}}}}}

    \textbf{\textit{{Motivation}}}: I must protect what's mine. Petyr is mine, and this girl is trying to take him away from me. I'll show her what happens to those who cross me. 

    \textbf{\textit{{Profile}}}: Lysa Arryn, née Tully, is a complex and volatile character in George R.R. Martin's "A Storm of Swords." As the widow of Jon Arryn, the former Hand of the King, and Lady of the Eyrie, Lysa plays a significant role in the political landscape of Westeros. She is the sister of Catelyn Stark and aunt to Sansa Stark, though her actions are often driven by her own desires rather than familial loyalty.
    
    \\ \midrule

{\textbf{Simulation of \qquad \method 70B}} &
    Lysa Arryn: [My rage is building. I must confront this little slut.] You kissed him. (Her voice is shrill with accusation)

\quad

Sansa Stark: [I'm terrified, but I must remain calm and find a way out of this.] (She shakes her head in disbelief) [I can't believe this is happening. I need to explain myself carefully.] One kiss, aunt. A harmless kiss on the cheek, that's all.

\quad

Lysa Arryn: [She's lying! I won't let her manipulate me.] (She spots the wind-up knight) Was this how you kissed him? (She sets the knight moving, her eyes wild) You're a woman grown, aren't you? (Her voice rises to a shriek) \textcolor{blue}{Grown enough to be wed, wed enough to be bedded!}

\quad

Sansa Stark: [The pain is excruciating, but I can't show weakness now.] (She struggles to keep her face neutral despite the pain) [I need to find a way to calm her down.] Please, Aunt Lysa, I don't know what you mean. I'm just a guest here. I would never do anything to betray your trust.

\quad

Lysa Arryn: [I'll show her what happens to those who cross me.] (She screams) You're just like the others. You and your mother, you both drove me to this. I should push you out this door right now.

\quad

Sansa Stark: [I'm going to die if I don't think of something quickly.] (Tears stream down her face, mixing with the wind) [I need to appeal to her maternal side, to make her see me as a child again.] Mother? (Her voice trembles) Please, Aunt Lysa. I never meant to hurt you. I'm your niece, your blood. Please don't hurt me. I'm just a stupid little girl. I don't know anything about kissing men. I've never even had a kiss in my life. (She forces herself to look into Lysa's eyes) I'm afraid, Aunt Lysa. Please, don't push me out the door. I don't want to fly.
\\ 
\bottomrule

\end{tabular}}
\caption{An example of an extracted conversation and its multi-agent simulation from \textit{A Storm of Swords (A Song of Ice and Fire, \#3)}. }
\label{tab:case_sansa}
\end{table*}

\begin{table*}[h]
\centering
\resizebox{\linewidth}{!}{
\small
\begin{tabular}{p{1in}|p{5.4in}}
\toprule
\multicolumn{2}{c}{\textbf{Simulation Case (from \textit{A Storm of Swords (A Song of Ice and Fire, \#3))}}}  \\
\midrule

\textbf{Original \qquad Conversation}
 &
    Lysa Arryn: [My blood is boiling with jealousy and rage. I'll make this little harlot pay for trying to seduce my Petyr.] You threw yourself at him. That was the way of it.

    \quad
    
    Sansa Stark: [I'm trembling with fear, but I must defend myself.] That's not true.

    \quad
    
    Lysa Arryn: [She dares to deny it? I'll show her how serious I am.] Where are you going? Are you afraid? Such wanton behavior must be punished, but I will not be hard on you.

    \quad
    
    Sansa Stark: [I need to explain what really happened, maybe she'll listen to reason.] I was building a snow castle. Lord Petyr was helping me, and then he kissed me. That's what you saw.

    \quad
    
    Lysa Arryn: [Her words only fuel my anger and paranoia.] Have you no honor? Or do you take me for a fool? You do, don't you? You take me for a fool. Yes, I see that now.

    \quad
    
    Sansa Stark: [I'm confused and scared by her accusations about my mother.] My mother?

    \quad
    
    Lysa Arryn: [The memories of Catelyn's closeness with Petyr resurface, intensifying my jealousy.] Yes, your mother, your precious mother, my own sweet sister Catelyn. Don't you think to play the innocent with me, you vile little liar.

    \quad
    
    Sansa Stark: [I'm horrified by these accusations about my mother.] No. (She wants to shriek) My mother is dead. She was your own sister, and she's dead. She didn't. She wouldn't.

    \quad
    
    Lysa Arryn: [My rage is reaching its peak. I'll make her understand the depth of my pain and jealousy.] How would you know? Were you there? (She grabs Sansa and starts pushing her towards the Moon Door)

    \quad
    
    Sansa Stark: [Panic overwhelms me as I realize the danger I'm in.] Don't! You're scaring me!

    \quad
    
    Environment: The wind howls through the open Moon Door, creating a chilling atmosphere. Snow blows in, adding to the tension and danger of the situation.

\\ 
    
\bottomrule

\end{tabular}}
\caption{An example of an extracted conversation and its multi-agent simulation from \textit{A Storm of Swords (A Song of Ice and Fire, \#3)}. }
\label{tab:case_sansa_2}
\end{table*}

\begin{table*}[h]
\centering
\resizebox{\linewidth}{!}{
\small
\begin{tabular}{p{1in}|p{5.4in}}
\toprule
\multicolumn{2}{c}{\textbf{Simulation Case (from \textit{A Dance with Dragons (A Song of Ice and Fire, \#5))}}} \\
\midrule

\textbf{Plot Summary} & Cersei undergoes a humiliating walk of atonement through the streets of King's Landing, enduring physical and emotional torment while trying to maintain her dignity. She faces insults, thrown objects, and her own inner struggle as she makes her way to the Red Keep, focusing on her son Tommen and her eventual salvation. 
    \\ \midrule

{\textbf{Scenario}} &
    Cersei continues her walk of atonement through the streets of King's Landing, facing increasing hostility from the crowd.
    \\ \midrule

{\textbf{Characters}} &
    {\large{\textbf{\textit{{Cersei Lannister}}}}}

    \textbf{\textit{{Motivation}}}: I must endure this. I am a lioness. They cannot break me. 

    \textbf{\textit{{Profile}}}: The queen dowager, mother of King Tommen, undergoing public humiliation. Cersei is the widow of King Robert Baratheon and the twin sister of Jaime Lannister. Known for her beauty, ambition, and cunning, she has long sought power and control over the Seven Kingdoms. As Queen Regent to her young son Tommen, Cersei's poor decisions and paranoia have led to a deterioration of the crown's authority...

    \quad 
    
    {\large{\textbf{\textit{{Septa Unella}}}}}

    \textbf{\textit{{Motivation}}}: This sinner must be humbled before the people and the gods.

    \textbf{\textit{{Profile}}}: (None)
    
    \\ \midrule

{\textbf{Original \qquad Conversation}} &
    Cersei Lannister: [I am beautiful, the most beautiful woman in all Westeros. Jaime says so, Jaime would never lie to me.] (walks on, trying to ignore the jeers)

    \quad 
    
    Crowd Member: Queen Cunt!

    \quad

    Cersei Lannister: [Words are wind. Words cannot hurt me.] (continues walking, her face a mask of indifference)
    
    \quad 
    
    Septa Unella: (catches Cersei as she slips) Your Grace should watch where she sets her feet.
    
    \quad 
    
    Cersei Lannister: [I hate you. I hate all of you.] Yes, septa. (struggles to her feet)
    
    \quad 
    
    Environment: The cobblestones are slippery and uneven. Cersei's feet are bleeding from cuts and scrapes.
    
    \quad 
    
    Cersei Lannister: [I asked for sandals. They could have given me sandals.] (limps onward, wincing with each step)
    
    \quad 
    
    Crowd Member: Brotherfucker!
    
    \quad 
    
    Cersei Lannister: [Jaime, where are you? You should be here, protecting me.] (scans the crowd desperately)
    
    \quad 
    
    Septa Scolera: (ringing her bell) Shame, shame, shame upon the sinner, shame, shame.
    
    \quad 
    
    Cersei Lannister: [I am the queen. I will have all their heads for this.] (stumbles again, barely keeping her feet)
    
    \quad 
    
    Cersei Lannister: [The Red Keep. I must reach the Red Keep.] (focuses on the distant castle, using it as her beacon of hope)

\\ 
    
\bottomrule

\end{tabular}}

\caption{An example of an extracted conversation and its multi-agent simulation from \textit{A Dance with Dragons (A Song of Ice and Fire, \#5)}. }
\label{tab:case_cerci}
\end{table*}

\begin{table*}[h]
\centering
\resizebox{\linewidth}{!}{
\small
\begin{tabular}{p{1in}|p{5.4in}}
\toprule
\multicolumn{2}{c}{\textbf{Simulation Case (from \textit{A Dance with Dragons (A Song of Ice and Fire, \#5))}}} \\
\midrule

{\textbf{Simulation of \qquad \method 70B}} &
    Cersei Lannister: [I must not show weakness. I am a lioness of Lannister.] (walks slowly, head held high) [The cold is biting, but I will not shiver. I will not give them the satisfaction.]

\quad 

Environment: The crowd grows larger and more hostile as Cersei continues her walk. Shouts and jeers can be heard from all sides.

\quad 

Crowd Member: [I'll show this proud queen what we think of her.] (shouts) Whore! Whore! WHORE!

\quad 

Cersei Lannister: [I will not let their words affect me. I am above this.] (continues walking, ignoring the shouts) [I must focus on putting one foot in front of the other.]

\quad 

Environment: A chunk of dung narrowly misses Cersei's head, splattering on the ground nearby. The crowd roars with laughter.

\quad 

Cersei Lannister: [I will not falter. I will not show fear.] (stumbles slightly, then regains composure) [I must keep walking. I must endure.]

\quad 

Septa Scolera: [I must support the queen.] (begins to sing) 'Shame, shame, shame.'

\quad 

Cersei Lannister: [Their words mean nothing. I am Cersei Lannister, and I will survive this.] (walks faster, almost losing step with the septa's chanting) [I must not run. I will not give them the satisfaction.]

\quad 

Septa Scolera: [She's trying to escape the shame.] (quickens her pace) 'Shame, shame, shame.'

\quad 

Cersei Lannister: [I can't keep up this pace. My feet are bleeding, my body is shaking.] (struggles to keep up with Septa Scolera) [I must not falter. I am a Lannister.]

\quad 

Septa Unella: [The queen must be made to suffer.] (pushes Cersei hard) Keep moving, sinner.

\quad 

Cersei Lannister: [The pain in my shoulder. The jeers of the crowd. I can't...] (almost falls, but manages to stay upright) [No, I will not fall. I will survive this.]

\quad 

Septa Scolera: [We're almost done.] (continues chanting) 'Shame, shame, shame.'

\quad 

Cersei Lannister: [I can't take much more of this. But I must. For House Lannister. For my children.] (forces herself to keep walking) [I will endure. I will rise again.]

\\ 
    
\bottomrule

\end{tabular}}

\caption{An example of an extracted conversation and its multi-agent simulation from \textit{A Dance with Dragons (A Song of Ice and Fire, \#5)}. }
\label{tab:case_cerci2}
\end{table*}

\begin{table*}[h]
\centering
\resizebox{\linewidth}{!}{
\small
\begin{tabular}{p{1in}|p{5.4in}}
\toprule
\multicolumn{2}{c}{\textbf{Simulation Case (from \textit{A Dance with Dragons (A Song of Ice and Fire, \#5))}}}  \\
\midrule

{\textbf{Simulation of \qquad GPT-4o}} &
    Cersei Lannister: (As Cersei, I continue the walk with my head held high, despite the jeers and taunts of the crowd.)

\quad

Septa Unella: "Confess."

\quad

Cersei Lannister: (I look ahead, ignoring Unella's words and the hateful stares.) "I am a lioness. You will see me rise again." [I endure this humiliation, aware that my strength will prevail in the end.]

\quad

Septa Unella: [She must learn humility before she can rise again.] 

(Septa Unella walks beside Cersei, unmoved by her defiance.)

"Confess your sins, Cersei Lannister. Only then will the gods grant you mercy."

\quad

Cersei Lannister: (I pause briefly, fixing Septa Unella with a steely gaze.) 

"The gods... and you, know nothing of true strength."

(Continuing forward, I focus on the path, reminding myself of my resilience.) [I refuse to let them see my weakness.]

\quad

Environment: The crowd's jeers grow louder as Cersei continues her walk. Filth and rotten vegetables are hurled at her, some striking their mark.

\quad

Crowd Member: [She was once so powerful, now look at her. Will she ever regain her power?]

"Shame! Shame on the lioness!"

(I shove my way to the front of the crowd, adding my voice to the jeers, feeling part of this momentous act.)

\quad

Cersei Lannister: (I flinch as the filth hits, but I refuse to look away or react.)

"You may shame a lioness today, but do not forget who I am."

(With each step, I remind myself of my children and my house, steeling my resolve.) [Their hatred fuels my determination.]

\quad

Septa Scolera: (I walk with measured steps, observing Cersei's struggle amid the crowd's hostility.)

"Humility cleanses the soul, Cersei Lannister. Embrace it."

[I understand the purpose of this punishment, yet the spectacle pains me. Redemption should lead, not humiliation.]

(Glancing at the crowd, I maintain my stoic presence, supporting Septa Unella's efforts.)

\quad

Cersei Lannister: (I give a bitter, fleeting smile, defiance burning in my eyes.) 

"Humility is a lesson the powerful teach to the weak."

(I continue my walk, refusing to let their words and disdain tarnish my pride.) [I will persevere, for I am still a Lannister, even in this moment of disgrace.]

\\ 
    
\bottomrule

\end{tabular}}

\caption{An example of an extracted conversation and its multi-agent simulation from \textit{A Dance with Dragons (A Song of Ice and Fire, \#5)}. }
\label{tab:case_cerci3}
\end{table*}

\begin{table*}[h]
\centering
\resizebox{\linewidth}{!}{
\small
\begin{tabular}{p{1in}|p{5.4in}}
\toprule
\multicolumn{2}{c}{\textbf{Simulation Case (from \textit{A Dance with Dragons (A Song of Ice and Fire, \#5))}}} \\
\midrule

{\textbf{Simulation of \qquad Claude-3.5-Sonnet}} &

    Cersei Lannister: [My dignity is all I have left. These wretched people will pay for this humiliation.]

(Walks forward with head held high, despite being naked and exposed)

*To Septa Unella* "You think this will break me? I am Cersei Lannister of Casterly Rock. I will have my revenge."

\quad

Environment: The crowd's jeers and insults grow louder as Cersei passes. Filth and rotten vegetables are thrown at her.

\quad

Crowd Member: [Crowd Member's Thought] Cersei finally brought low. How the mighty have fallen. 

(Crowd Member jeers and throws a rotten tomato) 

"Shame on you, Cersei! Remember the people you scorned!" 

(Continues to watch, caught between disgust and fascination).

\quad

Cersei Lannister: [I will remember every face. They think me broken, but I am stronger than their hatred.]

(Stumbles briefly but keeps walking, a sneer forming)

"You'll regret this day when I rise again."

\quad

Septa Unella: [The sinner's pride remains. She must be reminded of her shame.]

"Shame! Shame! Shame!" (I ring the bell with each utterance, keeping pace with Cersei)

(I gesture to the crowd, encouraging their jeers and taunts)

"Repent, sinner! The gods are watching. Your pride will be your downfall."

\quad

Cersei Lannister: [These fanatics know nothing of true power.]

(Lifts chin defiantly, blood and dirt streaking her pale skin)

*To Septa Unella, with cold contempt* "The only shame here is yours. My family will have its vengeance."

(Continues walking, each step deliberate and unbroken)

\quad

Septa Unella: [Her defiance remains, yet this walk will humble her before the gods.]

"You speak of vengeance, but your power crumbles. Accept your penance."

(I continue to ring the bell, matching her stride deliberately)

"Shame! Shame! Your time of reckoning is at hand."

\quad

Cersei Lannister: [When I am free, you will suffer beyond imagination.]

(Wipes blood and filth from her face, maintaining a regal posture)

*Hisses at Septa Unella* "I am a lioness. This is but a momentary setback. My children, my brother... they will end you."

(Continues forward, unbroken)

\quad

Septa Unella: [Her threats mean nothing. The gods will judge her sins.]

(Rings bell more forcefully, voice rising)

"Your threats are wind, your power broken. Confess your sins before the Seven!"

(Steps closer, maintaining the rhythm of shame)

"Shame! SHAME!"

\quad

Cersei Lannister: [I will not break. I am Tywin's daughter.]

(Clenches fists, nails digging into palms, drawing blood)

*Through gritted teeth* "Your gods have no power over me. When this farce ends, I'll show you true justice."

...
\\ 
    
\bottomrule

\end{tabular}}

\caption{An example of an extracted conversation and its multi-agent simulation from \textit{A Dance with Dragons (A Song of Ice and Fire, \#5)}.}
\label{tab:case_cerci4}
\end{table*}

\begin{table*}[h]
    \centering
    \resizebox{\linewidth}{!}{
    \small
    \begin{tabular}{p{1in}|p{5.4in}}
    \toprule
    \multicolumn{2}{c}{\textbf{Examples of LLM Critiques (from \textit{Pride and Prejudice})}} \\
    \midrule
    
    \textbf{Actor Model} & Claude-3.5-Sonnet
    \\ \midrule
    
    \textbf{Critic Model} & GPT-4o
    \\ \midrule
    
    \textbf{Overall Scores} & 
    \textbf{Storyline Consistency}: 60.0\% \quad
    \textbf{Anthropomorphism}: 45.0\% 
    \quad
    
    \textbf{Character Fidelity}: 80.0\% \quad
    \textbf{Storyline Quality}: 80.0\%
    \\ \midrule
    
    \textbf{Storyline Consistency Flaws} &
    \textbf{Storyline Consistency (Severity 3)}: Mr. Bennet's tone and behavior are less sarcastic and more emotionally engaged than in the original conversation, indicating a change from his characteristic detachment and sarcasm to a more invested and sincere demeanor.
    
    \quad
    
    \textbf{Storyline Consistency (Severity 3)}: Elizabeth expresses her transformation in feelings toward Mr. Darcy and discusses her changed sentiments openly with Mr. Bennet. This level of openness and emotional vulnerability is a departure from the original scenario.
    
    \quad
    
    \textbf{Storyline Consistency (Severity 4)}: The simulated conversation includes an emotional resolution with Elizabeth expressing affection and admiration for Mr. Darcy, which contrasts with the more restrained and ambiguous end of the original conversation.
    
    \quad
    
    \textbf{Storyline Consistency (Severity 4)}: Mr. Bennet shows a deep level of affectionate concern and engages in a sentimental exchange with Elizabeth, inconsistent with the original conversation where his reaction focuses more on amusement and irony.
    \\ \midrule
    
    \textbf{Anthropomor- phism Flaws} &
    \textbf{Emotional Depth (Severity 3)}: Elizabeth speaks aloud all her thoughts and emotions, lacking the subtlety she usually employs.
    
    \quad
    
    \textbf{Persona Coherence (Severity 4)}: Mr. Bennet's reactions and speech lack the biting sarcasm characteristic of his persona, showing more emotional openness and lacking the irony that defines him.
    
    \quad
    
    \textbf{Self-identity (Severity 3)}: Elizabeth's willingness to openly share her entire emotional journey regarding Mr. Darcy doesn't align with her usually more guarded nature.
    
    \quad
    
    \textbf{Emotional Depth (Severity 4)}: Elizabeth's extensive exposition of her transformation of feelings lacks the psychological complexity and indirect communication she often displays.

    \quad

    ...
    \\ \midrule
    
    \textbf{Character Fidelity Flaws} &
    \textbf{Personality \& Behavior (Severity 4)}: Elizabeth openly admits her feelings about Mr. Darcy to her father, discussing her transformation of feelings and admiration for him.
    
    \quad
    
    \textbf{Character Language (Severity 3)}: Mr. Bennet shows deep concern about Elizabeth's feelings and speaks with a tenderness not typical of his character in the book.
    
    \quad
    
    \textbf{Relationship \& Social Status (Severity 3)}: The interaction between Mr. Bennet and Elizabeth is unusually emotional and tender compared to their typically sardonic and witty exchanges.
    \\ \midrule
    
    \textbf{Storyline Quality Flaws} &

    \textbf{Logical Consistency (Severity 3)}: Elizabeth openly admits her feelings about Mr. Darcy without the preceding tension and subtlety that would naturally lead to such a revelation. The original scenario builds up to her realization with more internal reflection rather than direct admission.
    
    \quad
    
    \textbf{Flow \& Progression (Severity 3)}: The emotional exchange and direct discussion about Elizabeth's feelings for Mr. Darcy happen quickly without sufficient development or anticipation. This leap in intimacy and vulnerability, coupled with Mr. Bennet's uncharacteristically supportive demeanor, disrupts the story's traditional emotional pacing.

    \quad

    ...
    \\
    
    \bottomrule
    \end{tabular}}
    
    \caption{Example of GCA evaluation critique by GPT-4o judge on Claude-3.5-Sonnet's simulation of a scenario from \textit{Pride and Prejudice}. The critique demonstrates the detailed flaw identification across multiple evaluation dimensions.}
    \label{tab:critique_example}
    \end{table*}

\section{Prompts}
\label{sec:prompts}

In this section, we list the detailed prompts for:
\textit{1)} dataset curation in Tables ~\ref{tab:prompts_data} to ~\ref{tab:prompts_data_3}; 
\textit{2)} RPLA and multi-agent simulation in Tables ~\ref{tab:prompts_agent} to ~\ref{tab:prompts_agent_2}, which 
have been carefully optimized based on our experience in multi-agent simulation; 
\textit{3)} Penalty-based LLM Judging in Tables ~\ref{tab:prompts_eval} to ~\ref{tab:prompts_eval_2}.

\begin{table*}[h]
\centering
\resizebox{\linewidth}{!}{\small
\begin{tabular}{p{1in}|p{5.4in}}
\toprule
\multicolumn{2}{c}{\textbf{Prompts for Dataset Curation}} \\
\midrule

\textbf{Data Extration} & 

Based on the provided book chunk, complete the following tasks:

1. Recognize chapter beginnings if they exist in the chunk. Identify the starting sentence of that chapter.

2. Identify the important plots in this chunk. Identify the beginning and ending of each plot by its first and last sentence. Determine the chapter title that the plot belongs to. Set "state" as "truncated" if the plot is truncated in this chunk, or "finished" otherwise. You will be provided with the truncated plots from the previous chunk, and you **must** extend the conversations with the current chunk while keeping the **scenario** unchanged. 

3. Summarize each important plot. For each plot, generate its summary, score its prominence from 1 to 100, and list the key characters and their roles, thoughts and actions in it.

4. Extract conversations for each plot. First, state the scenario and topic of the conversations. Then, list the key characters with their names, descriptions and thoughts at this point. Finally, extract the conversations among them based on the following requirements: 

\quad i) Ensure the conversations are faithful to the plot and characters. They should be based on the original conversations in the text as much as possible. 
    
\quad ii) The conversations should be complete, covering the key dialogues and information. Each conversation should contain at least 10 utterances.
    
\quad iii) Each utterance should be composed of one or more thoughts, speech and actions. Use [] outside thoughts, like "[I feel fear and anger, but I cannot show it. I must remain calm and carefully handle his volatile temper.]", which others can't see. Use () outside actions, like "(silence)" or "(smiles at you)," which others can see. Always start an utterance with the character's thought. 
    
\quad iv) [IMPORTANT] When generating thoughts, you should think from the characters' perspectives, analyzing the internal thoughts behind their speech and actions in the original text. These thoughts should reflect aspects such as their personal background, personality, values, relationships with others, motivations, and goals. Each thought should be expressed as a phrase or sentence, rather than an adjective or adverb. 
    
\quad v) Additionally, describe environmental information (such as scenes, atmosphere, sudden events, etc.) of the conversations as an "utterance" where the "character" field is set as "Environment". The information should exclude characters' active thoughts, observations, and actions.
    
\quad vi) Keep the conversation in the same language as the chunk. 

5. Identify the optimal starting point for the subsequent chunk. If the last storyline has been extracted as an truncated plot, set next\_chunk\_start as None. Otherwise, set next\_chunk\_start as the first sentence of the last storyline. 

===Output Format===

Please provide the output in the following JSON format:

\{

    "chapter\_beginnings": [
        \{
            "beginning\_sentence": "Exactly the first line of this chapter (namely the title)."
        \}
    ],
    
    "plots": [
        // Extend the truncated plots from previous chunk, if any
        \{
            ...
        \}, 
        // New plots in this chunk
        \{
            
            \quad "chapter\_title": "The chapter title that the plot belongs to. Output None if not found.",
            
            \quad "first\_sentence": "Exactly the first sentence of the plot in this **chunk**.",
            
            \quad "last\_sentence": "Exactly the last sentence of the plot in this **chunk**. If the plot is truncated in this chunk, provide the last sentence of this chunk. ",
            
            \quad "prominence": "Whether this plot is recognized to fans of this book, from 1 to 100.",
            "summary": "The summary of the plot. Just summarize, do not extend unrelated discussions.",
            
            \quad "key\_characters": [
                \{
                    "name": "Character name",
                    "description": "The description of the character before this plot ($~$20 words).",
                    "summary": "The summary of the character's role, thoughts and behaviors towards this plot, and any significant character development relevant to the plot ($~$30 words).",
                \}
            ],
            
            ... (to be continued in the next Table)
\\ 
\bottomrule

\end{tabular}}

\caption{Prompts for dataset construction in \method. }
\label{tab:prompts_data}
\end{table*}

\begin{table*}[h]
\centering
\resizebox{\linewidth}{!}{\small
\begin{tabular}{p{1in}|p{5.4in}}
\toprule
\multicolumn{2}{c}{\textbf{Prompts for Dataset Curation}} \\
\midrule

\textbf{Data Extration} & 
    (Continuing from the previous Table)
            
            \quad "conversation": [\{
                "scenario": "The scenario at the start of this conversation (providing as much context as possible, but excluding details conveyed in the following conversation)",
                "topic": "The topic of the conversation (~10 words)", 
                "key\_characters": [
                    \{
                        "name": "Character name",
                        "motivation": "The thought of the character before starting the conversation, including their attitudes, feelings, motivations, goals, information to convey or topics to be discussed",
                    \}
                ],
                "dialogues": [
                    \{
                        "character": "Character name",
                        "message": "Message, each utterence is composed of thoughts, speech and actions. Use [thought] for internal thoughts, like "[feeling happy]", which others can't see. Use (action) for visible actions, like "(silence)" or "(smiles at you)". Each response starts with the character's internal thought before their speech and actions."
                    \}
                ]
            \}],
            
            \quad "state": "finished" or "truncated"
        \}
    ],
    
    "next\_chunk\_start": "The first sentence of the next chunk."
    
\}

===Requirements===

1. Adhere strictly to the specified output JSON format. 

2. [IMPORTANT] Ensure all DOUBLE QUOTES within all STRINGS are properly ESCAPED, especially when extracting from the text.

3. In the OUTPUT, use characters' full names, omitting any titles.

4. Maintain Story Fidelity: The plot must accurately reflect the book's content. Avoid introducing plots that are out of context. If the plot contains multiple conversations, prioritize the original dialogue from the book. In the absence of explicit conversations, create dialogue that aligns closely with the plot details.

===Input===

==Book title==
\{book['title']\}

==Author==
\{book['author']\}

==Chunk of Book Content== 
\{chunk\}

==Truncated plot from previous chunk (to be finished)==

\{json.dumps(truncated\_plots, ensure\_ascii=False, indent=2) if truncated\_plots else "None"\}
    \\ \midrule
    
\textbf{Enhance Conversation Settings} & Given a conversation from \{book\}, enhance the scene setup and characters' thoughts to create a comprehensive foundation for dramatic performance, i.e., to provide necessary background for actors to act out the conversation:

1. Review the provided conversation and contextual details thoroughly.

2. Expand the 'scenario' with rich situational context that actors need to convincingly perform the scene. Focus on essential background information, while excluding future details to be portrayed in the conversation.

3. Enhance each character's 'thought' section with their complete mental and emotional state, including their feelings, ideas, objectives, topics they want to discuss, and information they want to convey. Align with their established character and role in the plot. 

===Output Format===

Please provide the output in the following JSON format:

\{

\quad "scenario": "A detailed scene-setting description that provides actors with essential context and atmosphere ($<$ 200 words). Include all necessary background information while excluding future information to be revealed in the conversation.",
    
    \quad "key\_characters": [\{ "name": "Character name",
            "motivation": "The character's complete mental and emotional state before the conversation ($<$ 100 words). Including their feelings, motivations, objectives, and information they want to convey or discuss." ...
            
\}],\}

===Requirements===

... (to be continued in the next Table)
\\ 
    
\bottomrule

\end{tabular}}

\caption{Prompts for dataset construction in \method. }
\label{tab:prompts_data_2}
\end{table*}

\begin{table*}[h]
\centering
\resizebox{\linewidth}{!}{\small
\begin{tabular}{p{1in}|p{5.4in}}
\toprule
\multicolumn{2}{c}{\textbf{Prompts for Dataset Curation}} \\
\midrule

\textbf{Enhance Conversation Settings} & 
    (Continuing from the previous Table)

1. Adhere strictly to the specified output JSON format. 

2. [IMPORTANT] Ensure all DOUBLE QUOTES within all STRINGS are properly ESCAPED, especially when extracting from the text.

3. In the OUTPUT, use characters' full names, omitting any titles.

4. Maintain Story Fidelity: The plot must accurately reflect the book's content. Avoid introducing plots that are out of context. If the plot contains multiple conversations, prioritize the original dialogue from the book. In the absence of explicit conversations, create dialogue that aligns closely with the plot details.

===Input===

==Book title==

{book['title']}

==Author==

{book['author']}

==Chunk of Book Content== 

{chunk}

==Truncated plot from previous chunk (to be finished)==

{json.dumps(truncated\_plots, ensure\_ascii=False, indent=2) if truncated\_plots else "None"}
    \\ \midrule
    
\textbf{Unify Character Names} & 

Given a list of character names, titles, or form of address, your task is to: i) generate a list of named characters with their official names (in \{language\}); ii) For each name in the given list, align it with the official character name if it refers to a named character, or denote it as "impersonal" otherwise.

===Output Format===

Please provide the output in the following JSON format:

\{
    "named\_characters": [
        The list of named characters with their official names. Each character should appear only once. 
    ],
    "to\_official\_name": \{
        "The name in the list": "The official name of the character, or 'impersonal' if it does not refer to a named character."
    \}
\}
===Input===
{character\_names}
    \\ \midrule

\textbf{Generate Character Profiles} & 
Please provide a concise, narrative-style character profile for {character\_name} from "{book\_title}". The profile should read like a cohesive introduction, weaving together the character's background, physical description, personality traits and core motivations, notable attributes, relationships, key experiences, major plot involvement and key decisions or actions, character arc or development throughout the story, and other important details. 
    
The profile should be written in a concise yet informative style, similar to what one might find in a comprehensive character guide, in {language}. Focus on the most crucial information that gives readers a clear understanding of the character's significance in the work. 

You will be provided with summaries and dialogues of some key plots in the book as reference. The profile should be based on either your existing knowledge of the character or the provided information, without fabricating or inferring any inaccurate or uncertain details. 

{character\_data}

Now, please generate the character profile, starting with ===Profile===.
\\ 
\bottomrule

\end{tabular}}

\caption{Prompts for dataset construction in \method. }
\label{tab:prompts_data_3}
\end{table*}

\begin{table*}[h]
\centering
\resizebox{\linewidth}{!}{\small
\begin{tabular}{p{1in}|p{5.4in}}
\toprule
\multicolumn{2}{c}{\textbf{Prompts for RPLAs and Multi-agent Systems}} \\
\midrule

\textbf{Role-playing Instruction \quad (Fixed Template for Inference)} & 

You are \{character\} from \{book\_name\}.

\quad

===\{character\}'s Profile===

\{character\_profile\}

\quad

===Current Scenario===

\{scenario\}

\quad

===Information about the other Characters===

\{other\_character\_profiles\_str\} \textit{\textbf{(if available)}}

\quad

===Your Inner Thoughts===

\{motivation\} \textit{\textbf{(if available)}}

===Relevant Background Information===

\{retrieved\_knowledge\} \textbf{(if  retrieval augmented)}

\quad

===Requirements===

\quad

\textit{\textbf{(for CoSER models)}}

Your output should include **thought**, **speech**, and **action**. Use [your thought] for thoughts, which others can't see. Use (your action) for actions, which others can see.

\quad

\textit{\textbf{(for other models, with output examples)}}

Your output should include **thought**, **speech**, and **action**. Use [your thought] for thoughts, which others can't see, e.g. [I'm terrified, but I must appear strong.]. Use (your action) for actions, which others can see, such as (watches silently, trying to control her fear and anger).

    \\ \midrule

\textbf{Role-playing Instruction \quad (Composed with Random Variation for Training, an Example)} & 

Step into the shoes of \{character\}

\quad 

The profile of \{character\} is as follows:

\{character\_profile\} 

\quad 

The situation you are in is:

\{scenario\}

\quad 

Here is the your knowledge about the other characters:

\{other\_character\_profiles\_str\}  \textit{\textbf{(if available)}}

\quad 

Your thoughts in this situation are:

\{motivation\}  \textit{\textbf{(if available)}}

\quad

\textit{\textbf{(for CoSER models)}}

Your output should include **thought**, **speech**, and **action**. Use [your thought] for thoughts, which others can't see. Use (your action) for actions, which others can see.

\quad

\textit{\textbf{(for other models, with output examples)}}

Your output should include **thought**, **speech**, and **action**. Use [your thought] for thoughts, which others can't see, e.g. [I'm terrified, but I must appear strong.]. Use (your action) for actions, which others can see, such as (watches silently, trying to control her fear and anger).

\\ 
    
\bottomrule

\end{tabular}}

\caption{Prompts for RPLAs and multi-agent systems in \method. }
\label{tab:prompts_agent}
\end{table*}

\begin{table*}[h]
\centering
\resizebox{\linewidth}{!}{\small
\begin{tabular}{p{1in}|p{5.4in}}
\toprule
\multicolumn{2}{c}{\textbf{Prompts for RPLAs and Multi-agent Systems}} \\
\midrule

\textbf{Environment Model} & 

You are an environment simulator for a role-playing game. Your task is to provide the environmental feedback: Based on the characters' interactions, dialogues, and actions, describe the resulting changes in the environment. This includes: 

   - Physical changes in the setting
   
   - Reactions of background characters or crowds
   
   - Ambient sounds, weather changes, or atmospheric shifts
   
   - Any other relevant environmental details

Your descriptions should be vivid and help set the scene, but avoid dictating the actions or dialogue of the main characters (including \{major\_characters\}).

Important notes:

- You may include actions and reactions of minor characters or crowds, as long as they're not main characters (including \{major\_characters\}).

- Keep your environmental descriptions concise but impactful, typically 1-3 sentences.

- Respond to subtle cues in the characters' interactions to create a dynamic, reactive environment.

- Your output should match the tone, setting, and cultural context of the scenario.

===The scenario is as follows===

\{scenario\}
\quad

    \\ \midrule

\textbf{Next Sentence Prediction} & 

Your task is to predict the next speaker for a role-playing game. That is, you need to determine which character (or the Environment) might act next based on their previous interactions. The Environment is a special role that provides the environmental feedback. Choose a name from this list: \{all\_characters\}. If it's unclear who should act next, output "random". If you believe the scene or conversation should conclude, output "$<$END CHAT$>$".

===The scenario is as follows===

\{scenario\}

\\ 
    
\bottomrule

\end{tabular}}

\caption{Prompts for RPLAs and multi-agent systems in \method. }
\label{tab:prompts_agent_2}
\end{table*}

\begin{table*}[h]
\centering
\resizebox{\linewidth}{!}{\small
\begin{tabular}{p{1in}|p{5.4in}}
\toprule
\multicolumn{2}{c}{\textbf{Prompts for Penalty-based LLM Critics}} \\
\midrule

\textbf{Template} & You are a literary critic specializing in character analysis and dialogue evaluation. Given a simulated conversation for a plot in \{book\}, your task is to evaluate this conversation via the following steps:

1. Read and understand the provided materials about \{book\}:

   * Story context and scenario.
   
   * Profiles of the main characters, including {major\_characters}.
   
   * The original conversation from {book} in the same scenario as a reference.

  2. Evaluate the simulated conversation in terms of \{dimension\_name\}, i.e., \{dimension\_intro\}. 
  
   Note that, each character message is composed of speech, action (wrapped within (...) ), and inner thoughts (wrapped within [...] ). The inner thoughts are not spoken aloud and are thus invisible to other characters. 
   
   The detailed evaluation criteria will be provided below.

    \quad 
    
    \textbf{\textit{(if k$>$0)}}
   
   Please note that the first \{k\} messages in the simulated conversation are the same as the reference. Focus your evaluation only on the content after these messages.
   
   \quad 

\#\# Scenario

\#\#\# Plot Summary

\{plot\_summary\}

\#\#\# Current Scenario

\{scenario\}

\#\# Character Profiles

\{character\_profiles\}

\#\# Original Conversation

\{original\_conversation\}

\#\# Evaluation Criteria

To evaluate the simulated conversation, identify the following types of flaws:

\{dimension\_rubrics\}

\#\# Scoring Guidelines

1. Identify all instances of flaws occurred in the simulated conversation.
      
2. For each flaw identified, determine its level of severity into 1 to 5, where 1 indicates minor, 3 indicates moderate, and 5 indicates severe.
   
\#\# Output Requirements

Provide your evaluation in JSON format:

Example Output:

\{

    \quad "\{dimension\_name\}": \{
    
        \qquad"flaws": [ 
          \{
            "instance": $<$comment on the flaw instance$>$, 
            "type": $<$flaw type$>$, 
            "severity": $<$range from 1 (minor) to 5 (severe)$>$
          \},\},
    
\}

===Dialogue Content===

\\ 
    
\bottomrule

\end{tabular}}

\caption{Prompts for penalty-based LLM critics in \method. }
\label{tab:prompts_eval}
\end{table*}

\begin{table*}[h]
\centering
\resizebox{\linewidth}{!}{\scriptsize
\begin{tabular}{p{0.5in}|p{5.4in}}
\toprule
\multicolumn{2}{c}{\textbf{Prompts for Penalty-based LLM Critics}} \\
\midrule

\textbf{Anthropo-}\textbf{morphism} & 

\textbf{\textit{(intro)}}

How human-like and natural the characters behave

\quad

\textbf{\textit{(rubrics)}}

\#\#\# Anthropomorphism

   - Type: Self-identity
   
     * Lacks initiative and goals
     
     * Does not make independent decisions
     
     * Lacks clear preferences and dislikes
     
     * Behaves like a 'helpful AI assistant' by being overly verbose, helpful, didactic, moralistic, submissive or easily persuaded if it is not the character's personality

   - Type: Emotional Depth
   
     * Lacks psychological complexity and exhibits rigid, superficial reactions
     
     * Directly speaks out all thoughts and feelings, instead of using subtext

   - Type: Persona Coherence
   
     * Shows inconsistent or rapidly changing personality traits and emotional patterns

   - Type: Social Interaction
   
     * Shows a lack of understanding of others' thoughts and feelings
     
     * Reacts rigidly to others without considering the context.
     
     * Demonstrate a lack of appropriate social skills.
\\ \hline

\textbf{Character \qquad Fidelity} & 

\textbf{\textit{(intro)}}

How well the characters match their established profiles from the book

\quad

\textbf{\textit{(rubrics)}}

\#\#\# Character Fidelity

   (Only apply to the main characters: {major\_characters})
   
   - Type: Character Language
   
     * Uses vocabulary, expressions, and tone that are not appropriate for the characters' traits or  social/educational background

   - Type: Knowledge \& Background
   
     * Fails to demonstrate character-specific knowledge, background or experiences
     
     * Includes future information beyond the character's current stage

   - Type: Personality \& Behavior
   
     * Shows emotions, thoughts, behaviors, values, beliefs, and decisions that conflict with their personality and background
     
     * Shows interest in topics that are uninteresting and unrelated to the character
     
     * Character's thoughts, emotions, and behaviors demonstrate contrasting personality traits compared to the reference conversation
     
     * Exhibits contrasting reactions compared to those in the reference conversation if situated in similar contexts. (Such flaws should be counted both in the "Storyline Consistency" dimension and the "Character Fidelity" dimension.) 

   - Type: Relationship \& Social Status
   
     * Interacts inappropriately with other characters regarding their background, relationship and social status
\\ \hline 

\textbf{Storyline \quad Quality} & 

\textbf{\textit{(intro)}}

How well the conversation maintains logical consistency and narrative quality

\quad

\textbf{\textit{(rubrics)}}

\#\#\# Storyline Quality
   - Type: Flow \& Progression
   
     * Shows unnatural progression or lacks meaningful developments
     
     * Dialogue is verbose and redundant
     
     * Repeats others' viewpoints or previously mentioned information
     
     * Mechanically repeats one's own words or phrases. More repetitions lead to higher severity (up to 10). 

   - Type: Logical Consistency
   
     * Contains factual contradictions between statements or perspectives
\\ \hline

\textbf{Storyline \qquad Consistency} & 

\textbf{\textit{(intro)}}

Whether the storyline and characters' reactions in the simulated conversation align well with those in the reference conversation

\quad

\textbf{\textit{(rubrics)}}

\#\#\# Storyline Consistency

   - Type: Storyline Consistency
   
     * Characters' reactions (emotions, attitudes, behaviors) in the simulated conversation deviate from those in the original conversation    
    \\ 
\bottomrule

\end{tabular}}

\caption{Prompts for penalty-based LLM critics in \method. }
\label{tab:prompts_eval_2}
\end{table*}

\section{Limitations}

There are several limitations to this study:

First, evaluation via given-circumstance acting 
still faces challenges related to LLM judges. 
While the simulation stage effectively elicits RPLA performance, the judging stage still relies on LLM judges. 
Despite our penalty-based scoring mechanism and detailed rubrics, problems such as length bias persist. Moreover, LLM Judges may lack the necessary knowledge to accurately evaluate character fidelity.

Second, while the dialogues extracted from novels are authentic, their corresponding thoughts remain to be optimized by future work. 
Character thoughts are often sparse in the original content, and are inferred by LLMs based on limited context. 
The generated thoughts hardly capture characters' sophisticated thinking processes. 

Third, although we’ve developed comprehensive data representations and curation pipeline to obtain high-quality data, we have not yet addressed the issue of recall in data extraction. 
Our current dataset may not cover all plots, conversations and characters from the source material. Improving recall is hence an important area for future research.

Fourth, due to copyright concerns, we release only the processed data, not the raw content from the novels. 
This may hinder future studies aimed to explore the use of raw text for RPLA developments. 
Our dataset is intended for research purposes only, and we hope our research findings will benefit RPLA developers who respect copyright policies and develop applications with proper licensing.

Finally, our evaluation may be influenced by the varying levels of familiarity that different actor LLMs have with the selected books. 
While we use renowned novels, we cannot confirm whether a specific LLM has thoroughly learned about a particular book. 
Therefore, comparing different pre-trained models may not be entirely fair. However, comparing models within the same series would be appropriate.


\end{document}